\documentclass[a4wide,11pt]{article}
\usepackage{wrapfig,psfig,epsf}
\usepackage{natbib}

\usepackage{multicol}
\usepackage{xspace}
\usepackage{hyperref}
\usepackage{amssymb,amsmath}
\usepackage{multirow}
\usepackage[inline]{enumitem}
\usepackage{booktabs}
\usepackage{todonotes}
\usepackage{color}
\usepackage{xcolor}

\usepackage[nomessages]{fp}

\usepackage{subcaption}

\usepackage{graphicx}

\newcommand{\maxnum}{100.00}
\newlength{\maxlen}
\newcommand{\databar}[2][green!25]{%
  \settowidth{\maxlen}{\maxnum}%
  \addtolength{\maxlen}{\tabcolsep}%
  \FPeval\result{round(120*#2/\maxnum:4)}%
  \rlap{\color{green!25}\hspace*{-.5\tabcolsep}\rule[-.05\ht\strutbox]{\result\maxlen}{.95\ht\strutbox}}%
  \makebox[\dimexpr\maxlen-\tabcolsep][r]{#2}%
}

\newcommand{\databarred}[2][red!50]{%
  \settowidth{\maxlen}{\maxnum}%
  \addtolength{\maxlen}{\tabcolsep}%
  \FPeval\result{round(120*abs(#2)/\maxnum:4)}%
  \rlap{\color{red!50}\hspace*{-.5\tabcolsep}\rule[-.05\ht\strutbox]{\result\maxlen}{.95\ht\strutbox}}%
  \makebox[\dimexpr\maxlen-\tabcolsep][r]{#2}%
}

\definecolor{darkred}{rgb}{0.88, 0, 0}

\newcommand{\act}[1]{\ensuremath{\mathsf{#1}}}
\newcommand{\ppm}{Predictive Process Monitoring\xspace}

\newcommand{\bpm}{Business Process Monitoring\xspace}

\newcommand{\nirdi}{\texttt{Nirdizati}\xspace}
\newcommand{\HP} {Hyperparameter\xspace}

\newcommand{\hp} {hyperparameter\xspace}
\newcommand{\hps}{hyperparameters\xspace}
\newcommand{\Ds}{Dataset\xspace}
\newcommand{\Dss}{Datasets\xspace}
\newcommand{\ds}{dataset\xspace}
\newcommand{\dss}{datasets\xspace}

\newcommand{\M}{{\cal M}}
\newcommand{\Mzero}{{\cal M}_0}
\newcommand{\Mone}{{\cal M}_1}
\newcommand{\Mtwo}{{\cal M}_2}
\newcommand{\Mthree}{{\cal M}_3}
\newcommand{\Szero}{{\cal S}_0}
\newcommand{\Sone}{{\cal S}_1}
\newcommand{\Stwo}{{\cal S}_2}
\newcommand{\Sthree}{{\cal S}_3}
\newcommand{\tr}{{\cal TR}}
\newcommand{\trzero}{{\cal TR}_0}
\newcommand{\trone}{{\cal TR}_1}
\newcommand{\te}{{\cal TE}}

\newcommand{\rioone}{DriftRIO1\xspace}
\newcommand{\riotwo}{DriftRIO2\xspace}
\newcommand{\rio}{DriftRIO\xspace}

\newenvironment{scenario}[1][Scenario]{\begin{trivlist}
\item[\hskip \labelsep {\itshape #1}]}{\end{trivlist}}

\begin{document}

\title{How do I update my model? \\On the resilience of \ppm models to change}

\author{Williams Rizzi$^{1,2}$, Chiara {Di Francescomarino}$^1$, \\
 Chiara Ghidini$^1$,  Fabrizio Maria Maggi$^2$\\
$^1$ Fondazione Bruno Kessler (FBK), Trento, Italy \\
$^2$ Free University of Bozen-Bolzano, Bolzano, Italy \\
{\{wrizzi, dfmchiara, ghidini\}@fbk.eu}, {maggi@inf.unibz.it}
}

\date{}


 
\maketitle

\begin{abstract}
Existing well investigated \ppm techniques typically construct a predictive model based on past process executions, and then use this model to predict the future of new ongoing cases, without the possibility of updating it with new cases when they complete their execution.
This can make \ppm too rigid to deal with the variability of processes working in real environments that continuously evolve and/or exhibit new variant behaviours over time.
As a solution to this problem, we evaluate the use of three different strategies that allow the periodic rediscovery or incremental construction of the predictive model so as to exploit new available data. The evaluation focuses on the performance of the new learned predictive models, in terms of accuracy and time, against the original one, and uses a number of real and synthetic datasets with and without explicit Concept Drift.
The results provide an evidence of the potential of incremental learning algorithms for predicting process monitoring in real environments.

\end{abstract}

\section{Introduction}
\label{sec:intro}

\emph{\ppm}~\citep{DBLP:conf/caise/MaggiFDG14} is a research topic aiming at developing techniques that use the abundant availability of event logs extracted from information systems in order to predict how ongoing (uncompleted) process executions (a.k.a.\ cases) will unfold up to their completion. In turn, these techniques can be embedded within information systems to enhance their ability to manage business processes. For example, an information system can exploit a predictive monitoring technique to predict the remaining execution time of each ongoing case of a process~\citep{DBLP:conf/icsoc/Rogge-SoltiW13}, the next activity that will be executed in each case~\citep{DBLP:journals/dss/EvermannRF17}, or the final outcome of a case w.r.t.\ a set of possible outcomes~\citep{DBLP:conf/caise/MaggiFDG14, DBLP:conf/srii/MetzgerFE12, DBLP:journals/tsmc/MetzgerLISFCDP15}.

Existing \ppm techniques first construct a predictive model based on data coming from past process executions.
Then, they use this model to predict the future of an ongoing case (e.g., outcome, remaining time, or next activity).
However, when the predictive model has been constructed, it won't automatically take into account new cases when they complete their execution.
This is a limitation in the usage of predictive techniques in the area of \bpm: well-known characteristics of real processes are, in fact, their complexity, variability, and lack of steady-state.
Due to changing circumstances, processes (and thus their executions) evolve, increase their variability, and systems need to adapt in a timely manner.

While a rough answer to this problem would be the one of re-building new predictive models from the wider available set of data, one could observe that building predictive models has a cost and this option should therefore be well understood before embracing it; moreover, preliminary studies such as the one of~\citet{DBLP:conf/IEEEscc/MaisenbacherW17} investigate the usage of incremental techniques~\citep{DBLP:conf/esann/GepperthH16} in the presence of Concept Drift phenomena, thus suggesting a diverse strategy of updating a Predictive Process Monitoring model.

In this paper, we tackle the problem of updating Predictive Process Monitoring models in the presence of new process execution data in a principled manner by investigating, in a comparative and empirically driven manner, how different strategies to keep predictive models up-to-date work.
In particular, given an event log $\trzero$, and a set of new traces $\trone$, we focus on four diverse strategies to update a predictive model $\Mzero$ built using the traces of $\trzero$, to also take into account the set of new traces $\trone$:
\begin{itemize}
    \item \textbf{Do nothing}. In this case, $\Mzero$ is never updated and does not take into account $\trone$ in any way. This strategy acts also as a baseline against which to compare all the other strategies.
    \item \textbf{Re-train with no hyperopt}. In this case, a new predictive model $\Mone$ is built using $\trzero \cup \trone$ as train set but no optimisation of the \hps is performed and the ones of $\Mzero$ are used;
	\item \textbf{Full re-train}. In this case, a new predictive model $\Mtwo$ is built using $\trzero \cup \trone$ as the train set and a new optimisation of the \hps is performed;
	\item \textbf{Incremental update}. In this case, a new predictive model $\Mthree$ is built starting from $\Mzero$ using the cases contained in $\trone$ in an incremental manner (that is, using incremental learning algorithms).
\end{itemize}

The evaluation aims at investigating two main aspects of these update strategies: first, their impact on the quality\footnote{We measure the prediction quality using different metrics precisely defined later in the paper.} of the prediction in event logs that exhibit/do not exhibit Concept Drift; second, their impact on the time spent for building the predictive model $\Mzero$ and their updates.

The reason why we provide two different strategies for retraining the model, i.e., with no optimisation and with an optimisation of the \hps, is because the two costly activities when building a predictive model are the actual training of the model w.r.t. a train set and the optimisation of the \hps for the constructed model. Therefore, when evaluating the impact of retraining on an extended set of data we  aim at investigating the impact of building a new predictive model and the impact of optimising the \hp in a separate manner.

The problem upon which we investigate these four strategies is the one of outcome predictions, where the outcomes are expressed by using either Linear Temporal Logic (LTL) formulae \citep{pnueli1977temporal}, in line with several works such as~\citep{DBLP:journals/tsc/Francescomarino19,DBLP:conf/caise/MaggiFDG14}, or case duration properties. The four different strategies are evaluated in a broad experimental setting that considers different real and synthetic \dss.
Since we focus on outcome predictions, we have decided to center our evaluation on Random Forest. This algorithm was chosen as it was experimentally proven to be one of the best performing techniques on the outcome prediction problem - see \citep{DBLP:journals/tkdd/TeinemaaDRM19} for a rigorous review - and is therefore widely used on event log data usually used in \ppm.


Perhaps not surprisingly, the results show that the do-nothing strategy is not a viable strategy (and therefore the issue of updating a \ppm model is a real issue) and that full retraining and incremental updates are the best strategy in terms of quality of the updated predictive model. Nonetheless, the incremental update is able to keep up with the retraining strategy and deliver a properly fitted model almost in real time, whereas the full retraining might take hours and in some cases even days, suggesting that the potential of incremental models is under-appreciated, and clever solutions could be applied to deliver more stable performance while retaining the positive side of the \textit{update} functions.



The rest of the paper is structured as follows: Section~\ref{sec:background} provides the necessary background on \ppm and incremental learning;
Section~\ref{sec:running} presents two exemplifying scenarios of process variability and explicit Concept Drift; Section~\ref{sec:evaluation} illustrates the data and procedure we use to evaluate the proposed update strategies, while Section~\ref{sec:results} presents and discusses the results.
We finally provide some related work (Section~\ref{sec:related}) and concluding remarks (Section~\ref{sec:conclusions}).

\section{Background}
\label{sec:background}
In this section, we provide an overview of the four main building blocks that compose our research effort: \ppm , Random Forest, hyperparameter optimisation, and Concept Drift.

\subsection{\ppm}
\ppm~\citep{DBLP:conf/caise/MaggiFDG14} is a branch of Process Mining that aims at predicting at runtime and as early as possible the future development of ongoing cases of a process given their uncompleted traces. In the last few years, a wide literature about \ppm techniques has become available - see \citep{DBLP:conf/bpm/DiFrancescomarino18} for a survey - mostly based on Machine Learning techniques. The main dimension that is typically used to classify \ppm techniques is the type of prediction, which can belong to one of the three macro-categories: numeric predictions (e.g., time or cost predictions); categorical predictions (e.g., risk predictions or specific categorical outcome predictions such as the fulfillment of a certain property); next activities predictions (e.g, the sequence of the future activities, possibly with their attributes).

Frameworks such as \nirdi~\citep{DBLP:conf/bpm/RizziSFGKM19,DBLP:conf/bpm/JorbinaRVFDGMRR17} collect a set of Machine Learning techniques that can be instantiated and used for providing different types of predictions to the user.
In detail, these frameworks take as input a set of past executions and use them to train predictive models, which can then be stored to be used at runtime to continuously supply predictions to the user.
Moreover, the computed predictions can be used to compute accuracy scores for specific configurations.
Within these frameworks, we can identify two main modules: one for the \emph{case encoding}, and one for the \emph{supervised learning}. Each of them can be instantiated with different techniques.
Examples of case encodings are \emph{index-based encodings} presented in~\citet{DBLP:conf/bpm/LeontjevaCFDM15}.
Supervised learning techniques instead vary and can also depend on the type of prediction a user is interested in, ranging from Decision Tree and Random Forest, to regression methods and Recurrent Neural Networks.

\subsection{Random Forest}
Random Forest~\citep{DBLP:conf/icdar/Ho95} is an ensemble learning method used for classification and regression. The goal is to create a model composed of a multitude of Decision Trees~\citep{DBLP:journals/ml/Quinlan86}. In a Decision Tree, each tree interior node corresponds to one of the input variables and each leaf node to a possible classification or decision.
Each different path from the root to the leaf represents a different configuration of input variables. A tree can be ``learned'' by bootstrapping the source set into subsets based on an attribute value test. This process is repeated on each derived subset in a recursive manner called recursive partitioning.
The result is a tree in which each selected variable will contribute to the labelling of the relative example.
When an example needs to be labelled, it is run through all the Decision Trees. The output of each Decision Tree is counted. The most occurring label will be the output of the model.

In this work, we use non-incremental and incremental versions of Random Forest. In a nutshell, the non-incremental version builds a predictive model once and for all using a specific set of training data in a single training phase. Instead, in addition to the step of building a predictive model during the training phase, the incremental learning versions are able to update such a model whenever needed through an update function. The specific implementation used in this work incrementally updates the starting model by adding new decision trees as soon as new data is available.

As already mentioned in the Introduction, Random Forest was chosen as it was experimentally proven to be one of the best performing techniques on the outcome prediction problem in \ppm. The interested reader is referred to~\citet{DBLP:journals/tkdd/TeinemaaDRM19} for a rigorous review.

\subsection{\HP Optimisation}
\label{ssec:hp}
Machine Learning techniques are known to use model parameters and hyperparameters. Model parameters are automatically learned during the training phase so as to fit the data. Instead, hyperparameters are set outside the training procedure and used for controlling how flexible the model is in fitting the data. While the values of hyperparameters can influence the performance of the predictive models in a relevant manner, their optimal values highly depend on the specific dataset under examination, thus making their setting rather burdensome.
To support and automatise this onerous but important task, several \hp optimisation techniques have been developed in the literature \citep{Bergstra-Hyperopt,DBLP:conf/nips/BergstraBBK11} also for \ppm models -- see e.g., \citep{DBLP:journals/tkdd/TeinemaaDRM19,DBLP:journals/is/Francescomarino18,DBLP:journals/datamine/TeinemaaDLM18}. While, in~\citep{DBLP:journals/tkdd/TeinemaaDRM19,DBLP:journals/is/Francescomarino18}, the Tree Parser Estimator (TPE) has been used for outcome-oriented Predictive Process Monitoring solutions, in~\citep{DBLP:journals/datamine/TeinemaaDLM18}, Random Search has been used for hyperparameter optimisation in a comparative analysis focusing on how stable outcome-oriented predictions are along time. Although \hp optimisation techniques for \ppm have shown their ability to identify accurate and reliable framework configurations, they are also an expensive task and we have hence decided to evaluate the role of  hyperparameter tuning in our update strategies.

\subsection{Concept Drift}
\label{ssec:conceptDrift}
In Machine Learning, \emph{Concept Drift} refers to a change over time, in unforeseen ways, of the statistical properties of a target variable a learned model is trying to predict. This drift is often due to changes in the target data w.r.t. the one that was used in the training phase. These changes are problematic as they cause the predictions to become less accurate as time passes. Depending on the type of change (e.g., gradual, recurring, or abrupt), different types of techniques have been proposed in the literature to detect and handle them~\citep{DBLP:journals/csur/GamaZBPB14, DBLP:conf/sbia/GamaMCR04, DBLP:journals/ml/WidmerK96, DBLP:journals/ml/SchlimmerG86}.

Business Processes are subject to change due to, for example, changes in their normative, or organisational context, and so are their executions.  Processes and executions can hence be subject to Concept Drifts, which may involve several process dimensions such as its control-flow dependencies and data handling. For instance, an organisational change might affect how a certain procedure is managed by employees in a Public Administration scenario (e.g., a further approval by a new manager is required for closing the procedure), or a normative change might affect either the way in which patients are managed in an emergency department (e.g., patients have to be tested for COVID-19 before they can be visited) or the age of the customers who are allowed to submit a loan request procedure in a bank process. The Concept Drift phenomenon has originated few works that focus on drift detection and localisation in procedural and in declarative business processes - see \citep{DBLP:journals/tnn/BoseAZP14,DBLP:conf/ida/CarmonaG12} and \citep{DBLP:conf/otm/MaggiBCS13}, respectively - as well as on attempts to deal with it in the context of \ppm~\citep{DBLP:conf/IEEEscc/MaisenbacherW17,DBLP:conf/bpm/PauwelsC21}.


\section{Two Descriptive Scenarios}
\label{sec:running}

We aim at assessing the benefits of incremental learning techniques in scenarios characterised by process variability and/or explicit Concept Drift phenomena.
In this section, we introduce two typical scenarios, which refer to some of the \dss used in the evaluation described in Section~\ref{sec:evaluation}.

\begin{scenario}[Scenario 1. Dealing with Process Variability.] Information systems a\-re
wi\-de\-ly used in healthcare and several scenarios of predictive analytics can be provided in this domain.
Indeed, the exploitation of predictive techniques in healthcare is described as one of the promising big data trends in this domain \citep{Bughin:2017aa, MUNOZGAMA2022103994}.

Despite some successful evaluation of \ppm techniques using healthcare data~\citep{DBLP:conf/caise/MaggiFDG14}, predictive monitoring needs to consider a well known feature of healthcare processes, that is, their variability~\citep{DBLP:journals/jbi/RojasMSC16}, i.e., the variety of different alternative paths characterizing the executions of a process.
Whether they refer to non-elective care (e.g., medical emergencies), or elective care (e.g., scheduled standard, routine and non-routine procedures), healthcare processes often exhibit characteristics of high variability and instability. For instance, the treatment processes related to different patients can be quite different due to allergies or comorbidities or other specific characteristics of a patient.
In fact, when attempting to discover process model from data related to these processes, they are often spaghetti-like, i.e., cumbersome models in which it is difficult to distill a stable procedure.
Moreover, small changes in the organisational structure (e.g., new personnel in charge of a task, unforeseen seasonal variations due to holidays or diseases) may originate subtle variability not detectable in terms of stable Concept Drifts, but nonetheless relevant in terms of predictive data analytics.

In such a complex environment, an important challenge concerns the emergence of new behaviours: regardless of how much data we consider, an environment highly dependent on the human factor is likely to exhibit new variants that may not be captured when stopping the training at a specific time. Similarly, some variants may become obsolete, thus making the forgetting of data equally important.

Thus, a way for adapting the predictions to these changes, and an investigation of which update strategies are especially suited to highly variable and realistic process executions would be of great impact.
\end{scenario}

\begin{scenario}[Scenario 2. Dealing with explicit Concept Drift.]
The presence of Concept Drift in business processes, due to, e.g., changes in the organisational structures, legal regulations, and technological infrastructures, has been acknowledged in the Process Mining manifesto \citep{ProcessMiningManifesto} and literature~\citep{DBLP:conf/otm/MaggiBCS13}, together with some preliminary studies on its relation with \ppm~\citep{DBLP:conf/esann/GepperthH16}.

Such a sudden and abrupt variation in the data provides a clear challenge to the process owners: they must be ready to cope with a \ppm model with degraded performance on the drifted data, or to perform an update that allows the \ppm technique to support both the non-drifted and the drifted trends of the data (as ongoing executions may still concern the non-drifted cases).

Similarly to the above, an investigation of which update strategies are especially suited to realistic process executions that exhibit an explicit Concept Drift would provide a concrete support for the maintenance of \ppm models.
\end{scenario}

\section{Update Strategies}
\label{sec:strategies}

\ppm provides a set of techniques that use the availability of execution traces (or cases) extracted from information systems in order to predict how ongoing (uncompleted) executions will unfold up to their completion.
\begin{figure}
  \centering
    \includegraphics[width=.9\textwidth]{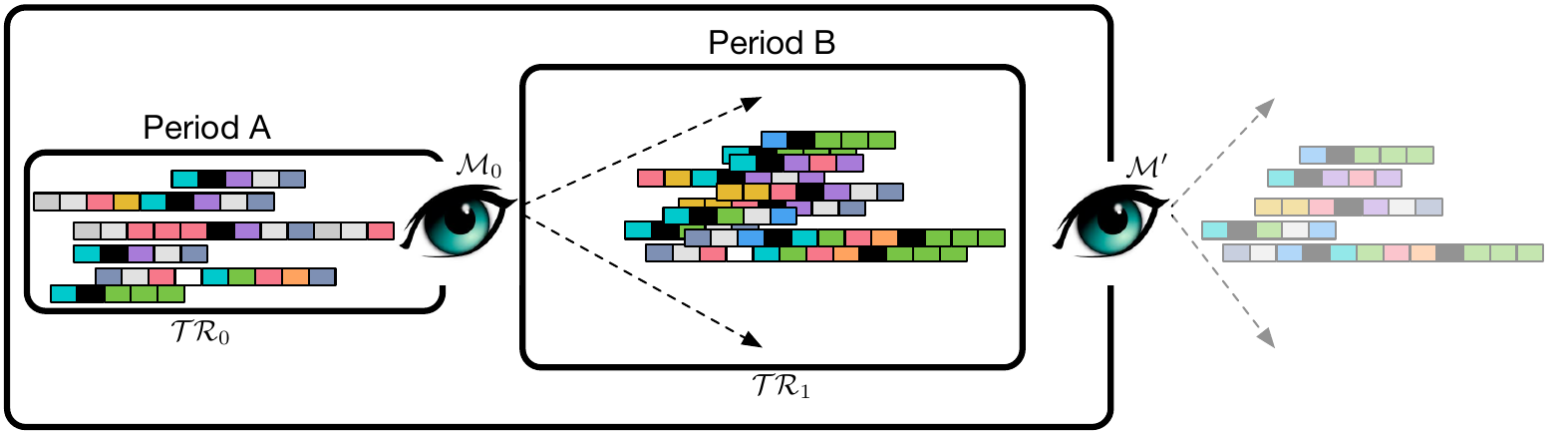}
  \caption{The general idea.}
  \label{fig:update-process}
\end{figure}
Thus, assume that an organisation was able to exploit a set of process executions $\trzero$, collected within a period of time that we will call ``Period A'' to obtain a predictive model $\Mzero$, and that it starts to exploit $\Mzero$ to perform predictions upon new incomplete traces (see Figure~\ref{fig:update-process} for an illustration of this scenario). As soon as the new incomplete traces terminate, they become new data, potentially available to be exploited for building a new model ${\cal M}'$, that, in turn can be used to provide predictions on new incomplete traces.
The need for exploiting new execution traces and building such an updated ${\cal M}'$ could be due to several reasons, among which the evolution of the process at hand (and thus of its executions) to which the system needs to adapt in a timely manner.

\begin{figure}
    \centering
      \includegraphics[width=.9\textwidth]{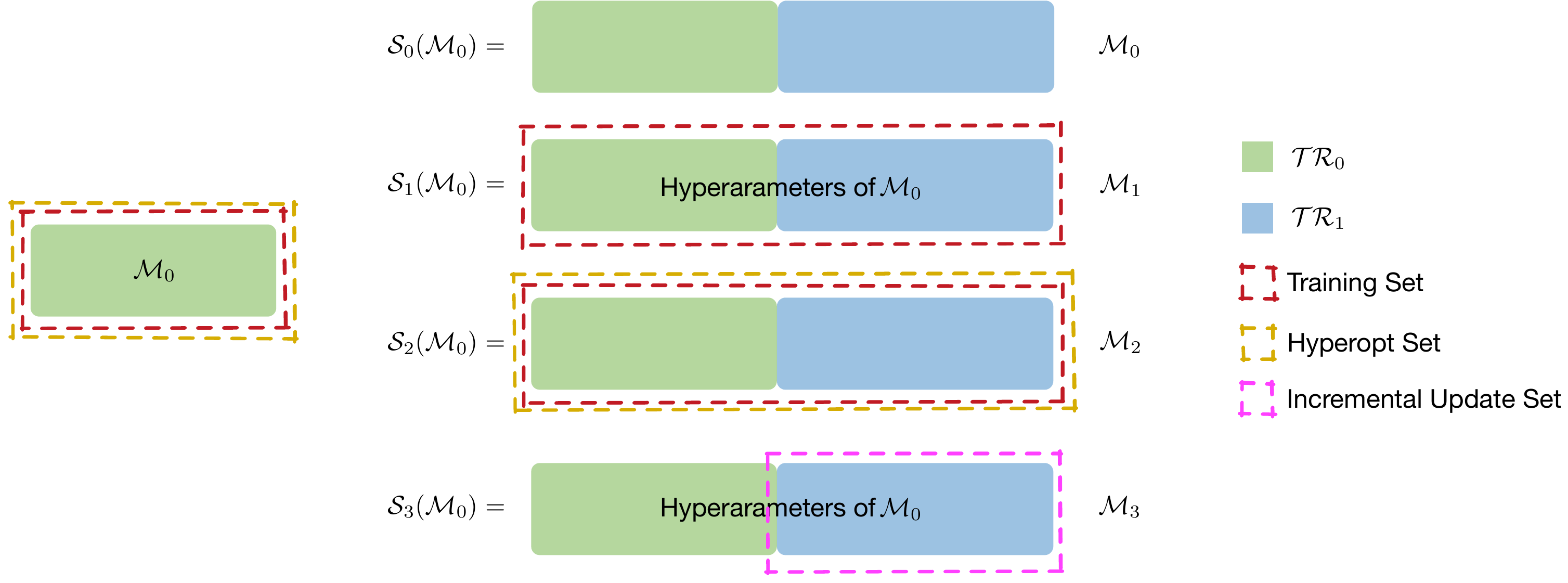}
    \caption{Four strategies to produce ${\cal M}'$.}
    \label{fig:figures_strategies}
  \end{figure}

In this paper, we provide 4 different strategies for computing ${\cal M}'$, by exploiting a new set of process executions $\trone$, collected in a period of time ``Period B'' subsequent to ``Period A'', along with the original set $\trzero$. The strategies are summarised in Figure~\ref{fig:figures_strategies}. The figure represents, on the left hand side, model $\Mzero$ and, on the right hand side, the operations performed starting from $\Mzero$ to obtain model ${\cal M}'$ according to the four update strategies.

The first strategy, $\Szero$, is a \textbf{do nothing} strategy. This strategy simply disregards that new traces are produced in ``Period B'' and continues to use $\Mzero$ as a predictive model. This strategy may prove to be useful when the processes remain stable and it acts also as a baseline against which to compare all the other strategies.

The second strategy, $\Sone$, exploits the new traces in $\trone$ produced in ``Period B'' for training but not for the optimisation of the hyperparameters. In this \textbf{re-train with no hyperopt} strategy, $\Mzero$ is replaced by a new predictive model $\Mone$ built from scratch by using $\trzero \cup \trone$ as train set. No optimisation of the hyperparameters is made in the construction of $\Mone$ and the values of the ones computed for $\Mzero$ are instead used. This strategy aims at exploiting the new data in $\trone$ still avoiding the costly steps needed for hyperparameter optimisation.

The third strategy, $\Stwo$, completely replaces the old model $\Mzero$ with a new predictive model $\Mtwo$ built from scratch using both $\trzero$ and $\trone$. This strategy aims at performing a \textbf{full re-train}, thus exploiting to the outmost all the available data. Also, the comparison between $\Sone$ and $\Stwo$ enables us to investigate the specific role of the hyperparameter tuning in the predictions.

The final strategy, $\Sthree$, exploits the new traces in $\trone$ produced in ``Period B'' for training the predictive model in an incremental manner. Differently from $\Sone$, the data of $\trone$ is added as training data in a continuous manner by means of incremental Machine Learning algorithms, to extend the existing knowledge of model $\Mzero$. The \textbf{incremental update} strategy is chosen as an example of dynamic technique, which can be applied when training data becomes available gradually over time or the size of the training data is too large to store or process it all at once. Similarly to $\Sone$, the value of the hyperparameters does not change when adding new training data.

\section{Empirical Evaluation}
\label{sec:evaluation}
The evaluation reported in this paper aims at understanding the characteristics of the four different update strategies introduced in the previous section in terms of accuracy and time. We aim at evaluating these strategies with two types of real-life event log data: event logs without an explicit Concept Drift and event logs with an explicit Concept Drift. As such, we have selected four real-life datasets\footnote{from the repository available at \url{https://data.4tu.nl/}}, two for the first scenario and two for the second one. To consolidate the evaluation on the Concept Drift scenario we also expanded the evaluation to include a synthetic event log with explicit Concept Drifts introduced in~\citet{DBLP:conf/bpm/MaaradjiDRO15}.

In this section, we introduce the research questions, the \dss, the metrics used to evaluate the effectiveness of the four update strategies described in Section~\ref{sec:strategies}, the procedure, and the tool settings. The results are instead reported in Section~\ref{sec:results}.

\subsection{Research Questions}
\label{ssec:rq}
Our evaluation is guided by the following research questions:
\begin{enumerate}[label=\textbf{RQ\arabic*.}, align=left]
	\item How do the four update strategies \textbf{do nothing}, \textbf{re-train with no hyperopt}, \textbf{full re-train}, and \textbf{incremental update} compare to one another in terms of accuracy?
	\item How do the four update strategies \textbf{do nothing}, \textbf{re-train with no hyperopt}, \textbf{full re-train}, and \textbf{incremental update} compare to one another in terms of time performance?
\end{enumerate}

\textbf{RQ1} aims at evaluating the quality of the predictions returned by the four update strategies, while \textbf{RQ2} investigates the time required to build the predictive models in the four scenarios and, in particular, aims at assessing the difference between the complete periodic rediscovery (\textbf{full re-train}) and the other two update strategies \textbf{re-train with no hyperopt} and \textbf{incremental update}.

\subsection{\Dss}
\label{ssec:dataset}
The four update strategies are evaluated using five \dss. Three of them are real-life event logs provided for a Business Process Intelligence (BPI) Challenge, in different years, without an explicit Concept Drift: the BPI Challenges 2011~\citep{BPIChallenge2011}, 2012~\citep{bpichallenge2012}, and 2015~\citep{bpichallenge2015}. They are examples of event logs exhibiting Process Variability as described in the first scenario in Section~\ref{sec:running}.
The remaining two datasets are instead examples of logs with explicit Concept Drift as described in the second scenario in Section~\ref{sec:running}. Our aim was to evaluate the four strategies  on real-life event logs, but, to the best of our knowledge, the only publicly available event log which contains an explicit concept drift is the BPI Challenge 2018~\citep{bpichallenge2018}. Therefore, we decided to augment the evaluation considering also one of the synthetic event logs introduced in \citet{DBLP:conf/bpm/MaaradjiDRO15}.
Here, we report the main characteristics of each \ds, while the outcomes to be predicted for each \ds are contained in Table \ref{table:PNs}.\footnote{We assume that events occurring during the process execution fall in the set of atomic propositions. LTL rules are constructed from these atoms by applying the temporal operators in addition to the usual boolean connectives. Given a formula $\varphi$, $\textbf{F} \varphi$ indicates that $\varphi$ is true sometimes in the future. $\textbf{G} \varphi$ means that $\varphi$ is true always in the future. $\varphi \textbf{U} \psi$ indicates that $\varphi$ has to hold at least until $\psi$  holds and $\psi$ must hold in the current or in a future time instant.}

\begin{table}
	\centering
    \scalebox{.88}{
\begin{tabular}{@{}ll@{}}
\toprule
\textbf{\Ds}  & \textbf{Outcome} \\
\midrule
\multirow{3}{*}{BPIC11} & {\scriptsize $\phi _{11} = \mathbf{G}(\act{CEA - tumor\ marker\ using\ meia} \rightarrow \mathbf{F}(\act{squamous\ cell\ carcinoma\ using\ eia}))$}\\
	& {\scriptsize $\phi _{12} = \neg(\act{histological\ examination - biopsies\ nno}) \mathbf{U} (\act{squamous\ cell\ carcinoma\ using\ eia})$}\\
	& {\scriptsize $\phi _{13} = \mathbf{F}(\act{histological\ examination} - \act{big\ resectiep})$}\\
\cmidrule{2-2}
\multirow{3}{*}{BPIC12} & {\scriptsize $\phi _{21} = \textbf{F}(\act{Accept\ Loan\ Application})$}\\
	& {\scriptsize $\phi _{22} = \textbf{F}(\act{Reject\ Loan\ Application})$}\\
	& {\scriptsize $\phi _{23} = \textbf{F}(\act{Cancel\ Loan\ Application})$}\\
	\cmidrule{2-2}
\multirow{3}{*}{BPIC15} & {\scriptsize $\phi _{31} = \textbf{F}(\act{start\ WABO\ procedure})\wedge \mathbf{F}(\act{extend\ procedure\ term})$} \\
& {\scriptsize $\phi _{32} = \textbf{F}(\act{receive\ additional\ information})\vee \mathbf{F}(\act{enrich\ decision})$} \\
& {\scriptsize $\phi _{33} = \textbf{G}(\act{send\ confirmation\ receipt} \rightarrow \mathbf{F}(\act{retrieve\ missing\ data}))$}\\
\cmidrule{2-2}
	BPIC18  & {\scriptsize $\phi _{41} = \textbf{F}(\act{Calculate \rightarrow (Parcel\ document \vee Geo\ parcel\ document)})\vee \textbf{F}(\act{Finish\ editing} \rightarrow \act{Begin\ editing})$}\\
		\cmidrule{2-2}
DriftRIO1 & {\scriptsize $\phi _{51} =$ fast case}\\ 		\cmidrule{2-2} 
DriftRIO2 & {\scriptsize $\phi _{61} =$ fast case}\\ 

\bottomrule
\end{tabular}
    }
\caption{The outcome formulae.}
\label{table:PNs}
\end{table}

The first \ds, originally provided for the BPI Challenge 2011, contains the treatment history of patients diagnosed with cancer in a Dutch academic hospital.
The log contains 1,140 cases and 149,730 events referring to 623 different activities.
Each case in this log records the events related to a particular patient. For instance, the first labelling ($\phi_{11}$), for this dataset, is such that the positive traces are all the ones for which if activity \act{CEA - tumor\ marker\ using\ meia} occurs, then it is followed by an occurrence of activity \act{squamous\ cell\ carcinoma\ using\ eia}.

The second \ds, originally provided for the BPI Challenge 2012, contains the execution history of a loan application process in a Dutch financial institution.
It is composed of 4,685 cases and 186,693 events referring to 36 different activities.
Each case in this log records the events related to a particular loan application.
For instance, the first labelling ($\phi_{21}$), for this dataset, is such that the positive traces are all the ones in which event \act{Accept\ Loan\ Application} occurs.

The third \ds, originally provided for the BPI Challenge 2015, concerns the application process for construction permits in five Dutch municipalities.
We consider the log pertaining to the first municipality, which is composed of 1,199 cases and 52,217 events referring to 398 different activities.

The fourth \ds, originally provided for the BPI Challenge 2018, concerns an event log from the European Agricultural Guarantee Fund pertaining to an application process for EU direct payments for German farmers from the European Agricultural Guarantee Fund. Depending on the document types, different branches of the workflow are performed. The event log used in this evaluation is composed of 29,302 cases and 1,661,656 events referring to 40 different activities.

The fifth and the sixth datasets, hereafter called \rioone and \riotwo, are synthetic event logs that use a ``textbook'' example of a business process for assessing loan applications~\citep{DBLP:journals/dke/WeberRR08}. The \rio event logs introduced in~\citet{DBLP:conf/bpm/MaaradjiDRO15} have been built by alternating traces executing the original ``base'' model and traces modified so as to exhibit complex Concept Drifts obtained by composing simple log changes, namely, re-sequentialisation of process model activities (R), insertion of a new activity (I) and optionalisation of one activity (O).\footnote{The logs used in~\citet{DBLP:conf/bpm/MaaradjiDRO15} have been slightly modified (i) by further strengthening the introduced Concept Drift with the increase of the time duration of two activities, as well as (ii) by considering only the first parts of the event logs, so as to have a single occurrence of the Concept Drift rather than multiple occurrences as in the original event logs.}
\rioone is composed of 3,994 cases and 47,776 events related to 19 activities. \riotwo is composed, instead, of 2,000 cases and 21,279 events referring to 19 different activities.
The outcomes to be predicted for each \ds are expressed by using LTL formulae for the four BPI Challenges~\citep{DBLP:journals/tsc/Francescomarino19,DBLP:conf/caise/MaggiFDG14} and by using the case duration property of being a fast case for \rio\footnote{Specifically, the property defines fast a case with a cycle time lower than the average cycle time of the event log.} (see Table \ref{table:PNs}).

\subsection{Metrics}
\label{ssec:metrics}
In order to answer the research questions, we use two metrics, one for accuracy and one for time.
The one for accuracy is used to evaluate \textbf{RQ1}, whereas the time measure is  used to evaluate \textbf{RQ2}.

\begin{description}
	\item[\textbf{The accuracy metric.}] In this work, we exploit a typical evaluation metric for calculating the performance of a classification model, that is AUC-ROC (hereafter only AUC). The ROC curve is a graphical plot that illustrates the diagnostic ability of a binary classifier system as its discrimination threshold is varied. The curve is created by plotting the true positive rate (TPR) against the false positive rate (FPR) using various threshold settings. In formulae, $\textit{TPR} = \frac{T_P}{T_P + F_N}$, and $\textit{FPR} = \frac{F_P}{FP + T_N}$, where $T_P$, $T_N$, $F_P$, and $F_N$ are the true-positives (positive outcomes correctly predicted), the true-negatives (negative outcomes correctly predicted), the false-positives (negative outcomes predicted as positive), and the false-negatives (positive outcomes predicted as negative), respectively.
		In our case, the AUC is the area under the ROC curve and, when using normalised units, it can be intuitively interpreted as the probability that a classifier will rank a randomly chosen positive instance higher than a randomly chosen negative one.
		As usual, $T_P$, $T_N$, $F_P$, and $F_N$ are obtained by comparing the predictions produced by the predictive models against a \emph{gold standard} that indicates the correct labelling of each case. In our experiments, we have built the gold standard by evaluating the outcome of each completed case in the test set.\footnote{To avoid biases related to the chosen metric, in our experiments we have also measured accuracy in terms of average F-measure and accuracy defined as $\frac{T_P + T_N}{T_P + T_N + F_P + F_N}$, and the results remain consistent. We have, therefore, decided to focus only on AUC for the sake of simplicity and readability of the results.}
	\item[\textbf{The time metric.}] We measure the time spent to build the predictive model in terms of execution time. The execution time indicates the time required to create and update (in the case of incremental algorithms) the predictive models. We remark here that the execution time does not include the time spent to load and pre-process the data, but only the bare processing time.
\end{description}

\subsection{Experimental Procedure}
\label{ssec:scenarios}

We adopt the classical Machine Learning Train/Validate/Test experimental procedure and configure it for the four different strategies we want to compare. The procedure consists of the following main steps:
\begin{enumerate*}[label=(\arabic*)]
    \item \ds preparation;
    \item classifier training and validation;
    \item classifier testing and metrics collection.
\end{enumerate*}

In the \ds preparation phase, the execution traces are first ordered according to their starting date-time so as to be able to meaningfully identify those referring to ``Period A'' (that is, $\trzero$), those referring to the subsequent ``Period B'' (that is, $\trone$), and those referring to the test set $\te$, which here represents the most recent set of traces where the actual predictions are made. The predictions are tested using the four different models $\Mzero$-$\Mthree$ corresponding to the different strategies $\Szero$-$\Sthree$ described in Section~\ref{sec:strategies}.

\begin{figure}
	\centering
		\includegraphics[width=.9\textwidth]{Splits.pdf}
	\caption{The experimental settings used to build the predictive models.}
	\label{fig:splits}
\end{figure}

The actual splits between train, \hp validation, and test sets used for evaluating the four strategies are illustrated in Figure~\ref{fig:splits}. Following a common practice in Machine Learning, we have decided to use 80\% of the data for training/validation and 20\% for testing.

\begin{table}
	\centering
    \scalebox{.75}{
		\begin{tabular}{@{}l c c c c c c@{}}
		\toprule
		\multirow{4}{*}{\textbf{Dataset}} & \multicolumn{3}{c}{\textbf{Trace Entropy}} & \multicolumn{3}{c}{\textbf{Global Block Entropy}} \\
		& $\mathbf{\tr \cup \te}$ & $\mathbf{\trzero \cup \te}$ & $\mathbf{\trone \cup \te}$ & $\mathbf{\tr \cup \te}$ & $\mathbf{\trzero \cup \te}$ & $\mathbf{\trone  \cup \te}$ \\
		& \multirow{2}{*}{\textbf{0\%-100\%}} & \textbf{0\%-40\% -} & \textbf{40\%-80\% -} & \multirow{2}{*}{\textbf{0\%-100\%}} & \textbf{0\%-40\% -} & \textbf{40\%-80\% -} \\
		&  & \textbf{80\%-100\%} & \textbf{80\%-100\%} &  & \textbf{80\%-100\%} & \textbf{80\%-100\%}  \\ \midrule
		BPIC11 & 9.62 & 9.08 & 8.97 & 24.71 & 23.73 & 24.13\\
		BPIC12 & 11.65 & 11.05 & 11.08 & 16.18 &  15.9 & 16.08\\
		BPIC15 & 10.16 & 9.4 & 9.48 & 18.96 & 18.55 & 18.4\\ 
		BPIC18 & 12.88 & 12.7 & 11.56 & 20.72 & 20.54 & 19.74\\ 
		DriftRIO1 & 4.66 & 4.71 & 4.29 & 8.52 & 8.54 & 8.38 \\ 
		DriftRIO2 & 4.27 & 4.29 & 4.1 & 8.41 & 8.41 & 8.37 \\
		\bottomrule	
		\end{tabular}
	}
	\caption{Dataset entropy.}
	\label{table:entropy}
\end{table} 

To test whether the performance of the different strategies was connected to different sizes of $\trzero$ and $\trone$, we devised two experimental settings to supply the train data to the learning algorithms.
The first experimental setting divides the train set unequally in $ \trzero $ (10\%) and $ \trone $ (70\%); the second experimental setting divides the train set equally in $\trzero$ (40\%) and $\trone$ (40\%). These two settings do not concern the construction of $\Mtwo$, which is always built using the entire set of available train data. The \hp validation set is extracted from the train set through a randomised sampling procedure. It amounts to 20\% of the train set, which corresponds to 2\% of the \ds for the train set at 10\%, 8\% for the train set at 40\%, and 16\% for the train set at 80\% (see Figure \ref{fig:splits}).

The variability of the log behaviour of each dataset can be measured through the \emph{trace entropy} and the \emph{Global block entropy} metrics~\citep{DBLP:journals/jodsn/BackDS19} (Table~\ref{table:entropy}). While the first metric mainly focuses on the number of variants in an event log, the latter also takes into account the internal structure of the traces. Note that the BPI Challenge 2018 and \rio datasets contain a Concept Drift in $\trone$ in both settings (it affects the last 30\% of data in $\tr$, when ordered according to their starting date-time) and, for these datasets, the difference between the entropy value of the split 0\%-40\%-80\%-100\% is higher than the entropy value of the split 40\%-80\%-80\%-100\% for both entropy metrics. Comparing the behaviour variability of the splits 0\%-10\%-80\%-100\% and 10\%-80\%-80\%-100\% is instead less useful since the two sets of traces have different sizes and the smaller set has obviously a lower entropy w.r.t.\ the larger one for all datasets. Finally, the entropy on the complete datasets (column 0\%-100\%) is useful to understand what type of input is provided when evaluating strategy $\Stwo$ that uses the full dataset to train the predictive model.

\begin{table}
	\centering
    \scalebox{.7}{
		\begin{tabular}{@{}l l c c c c c c c c c c@{}}
		\toprule
		\multirow{3}{*}{\textbf{Dataset}} & \multirow{3}{*}{\textbf{Formula}} & \multicolumn{4}{c}{$\mathbf{\trzero}$} & \multicolumn{2}{c}{$\mathbf{\tr}$} & \multicolumn{2}{c}{$\mathbf{\te}$} & \multicolumn{2}{c}{\textbf{Total}}\\
		& & \multicolumn{2}{c}{\textbf{0\%-10\%}} & \multicolumn{2}{c}{\textbf{0\%-40\%}} & \multicolumn{2}{c}{\textbf{0\%-80\%}} & \multicolumn{2}{c}{\textbf{80\%-100\%}} & \multicolumn{2}{c}{\textbf{0\%-100\%}}\\
		& & \textbf{True} & \textbf{False} & \textbf{True} & \textbf{False} & \textbf{True} & \textbf{False} & \textbf{True} & \textbf{False} & \textbf{True} & \textbf{False}\\ \midrule
		\multirow{3}{*}{BPIC11}   & $\phi _{11}$ & 50 & 64 & 156 & 300 & 359 & 663 & 109 & 119 & 458 & 682 \\ 
															& $\phi _{12}$ & 96 & 18 & 391 & 65 & 742 & 170 & 151 & 77 & 893 &247 \\ 
															& $\phi _{13}$ & 20 & 94 & 79 & 377 & 161 & 639 & 180 & 732 & 259	& 881\\ \midrule
		\multirow{3}{*}{BPIC12}   & $\phi _{21}$ & 227 & 241 & 904 & 970 & 1761 & 1987 & 482 & 455 & 2243 &	2442 \\
															& $\phi _{22}$ & 163 & 305 & 663 & 1211 &1366 & 2382 & 274 & 663 & 1640	& 3045\\ 
															& $\phi _{23}$ & 78 & 390 & 307 & 1567 & 621 & 3127 & 181 & 756 & 802	& 3883 \\ \midrule
		\multirow{3}{*}{BPIC15}   & $\phi _{31}$ & 6 & 90 & 7 & 472 & 59 & 900 & 63 & 178 & 122	& 1078 \\ 
															& $\phi _{32}$ & 33 & 63 & 110 & 369 & 223 & 736& 110 & 131 & 333	& 867\\ 
															& $\phi _{33}$ & 33 & 63 & 124 & 355 & 215 & 744 & 83 & 158 & 298	& 902\\ \midrule
		BPIC18   									& $\phi _{41}$ & 1994 &	936	& 7787 & 3933	& 16351 &	7090 & 4384	& 1477 & 20735 & 8567\\ \midrule
		DriftRIO1 								& $\phi _{51}$ & 25	& 272 &	132 &	1081 & 989 & 1987 &	582	& 436 & 1571 & 2423\\ \midrule
		DriftRIO2   							& $\phi _{61}$ & 83 &	117 & 398 &	402	& 834 &	766 &	229 &	171 & 1063 & 937\\
		\bottomrule	
		\end{tabular}
	}
	\caption{Dataset label distribution.}
	\label{table:labels}
\end{table} 

Additional information about the input logs used in our experiments is reported in Table~\ref{table:labels} that shows, for each dataset and for each setting, the distribution of the labels on both the train and the test sets, thus providing an idea of how much balanced the datasets are.

Once the data is prepared, training and validation start by extracting execution prefixes and by encoding them using the complex index encoding introduced by~\citet{DBLP:conf/bpm/LeontjevaCFDM15}.\footnote{We have decided to use the complex index encoding as it is the one providing better performance in \ppm~-- see~\citep{DBLP:conf/bpm/LeontjevaCFDM15}.} In the complex index encoding, the data related to a process execution is divided into static and dynamic information. Static information is the same for all the events in
the sequence (e.g., the age of a patient), while dynamic information changes for different events occurring in the process execution (e.g., the name of the activity, the pressure value of a patient associated to an event aimed at measuring the patient's pressure). We encode static information as attribute-value pairs, while, for dynamic information, the order in which events (and the related attributes) occur in the sequence is also taken into account.

The \hp optimisation function uses the Tree of Parzen Estimators (TPE) \footnote{TPE has been shown to be a good solution for complex hyperparameter optimisation problems~\citep{DBLP:conf/nips/BergstraBBK11} and is traditionally used for outcome-oriented Predictive Process Monitoring solutions~\citep{DBLP:journals/tkdd/TeinemaaDRM19}.} to retrieve the hyperparameter configuration (see Section~\ref{ssec:hp}), with a maximum of 1000 iterations, and the AUC as objective evaluation measure to maximise. Models $\Mzero$-$\Mthree$ are produced at the end of this phase. The specific hyperparameters used by the models are reported in Table~\ref{table:hyperparameters}. The table shows that indeed differences exist in the optimized hyperparameters computed starting from different train sets, i.e., 10\% of the datasets, 40\% of the datasets (both used for $\Mzero$, $\Mone$, and $\Mthree$) and 80\% of the datasets (used for $\Mtwo$).

\begin{table}
	\centering
    \scalebox{.6}{
		\begin{tabular}{@{}l l c c c@{\hskip 0.8cm} c c c@{\hskip 0.8cm} c c c@{}}
		\toprule
		\multirow{3}{*}{\textbf{Dataset}} & \multirow{3}{*}{\textbf{Formula}} & \multicolumn{3}{c}{\textbf{0\%-10\%}}  & \multicolumn{3}{c}{\textbf{0\%-40\%}} & \multicolumn{3}{c}{\textbf{0\%-80\%}}  \\
		& & \multicolumn{3}{c}{$\mathbf{\Mzero,\Mone,\Mthree}$} & \multicolumn{3}{c}{$\mathbf{\Mzero,\Mone,\Mthree}$} & \multicolumn{3}{c}{$\mathbf{\Mtwo}$} \\
		& & \textbf{\# est} & \textbf{max\_d} & \textbf{max\_f} & \textbf{\# est} & \textbf{max\_d} & \textbf{max\_f} & \textbf{\# est} & \textbf{max\_d} & \textbf{max\_f}  \\ \midrule
		\multirow{3}{*}{BPIC11}   &$\phi_{11}$ & 163 & 4  & log$_2$(n) & 537 & 8  & sqrt(n)   & 397 & 15 & auto     \\ 
															&$\phi_{12}$ & 273 & 9  & log$_2$(n) & 155 & 7  & auto      & 991 & 7	 & n         \\ 
															&$\phi_{13}$ & 395 & 28 & n          & 865 & 12 & n         & 203 & 7  & n        \\ \midrule
		\multirow{3}{*}{BPIC12}   &$\phi_{21}$ & 157 & 16 & sqrt(n)    & 427 & 14 & sqrt(n)   & 206 & 9	 & auto      \\ 
															&$\phi_{22}$ & 700 & 6	& auto	     & 558 &	9	& sqrt(n)   & 536 & 6  & auto      \\ 
															&$\phi_{23}$ & 180 & 7	& log$_2$(n) & 163 & 20 & auto	    & 587 & 27 & log$_2$(n) \\ \midrule
		\multirow{3}{*}{BPIC15}   &$\phi_{31}$ & 725 & 15 & log$_2$(n) & 377 & 19 & n	        & 563 & 16 & sqrt(n)   \\ 
															&$\phi_{32}$ & 977 & 13 & auto       & 937 & 8	& sqrt(n)   & 417 & 14	& sqrt(n)   \\ 
															&$\phi_{33}$ & 374 & 29 & auto   	   & 374 & 29 & auto      & 161 & 13 & sqrt(n)    \\ \midrule
		BPIC18 									&$\phi_{41}$ & 990 & 8	& n          & 650 & 26 & sqrt(n)   & 165 & 24	& n         \\ \midrule
		DriftRIO1									&$\phi_{51}$ & 913 & 4	& log$_2$(n) & 971 & 9	& n         &	293	& 4	  & log$_2$(n) \\ \midrule
		DriftRIO2									&$\phi_{61}$ & 998 & 23	& n          & 151 & 4  & sqrt(n)   & 178	& 5	  & auto       \\
		\bottomrule	
		\end{tabular}
	}
	\caption{Model hyperparameters.}
	\label{table:hyperparameters}
\end{table}

After the training and validation procedures are completed, we test the resulting model with the test set, we collect the scored metrics (see Section~\ref{ssec:metrics}), and store them in a database.
Concerning time, we have decided to measure here the time spent to build the initial  model $\Mzero$, plus the time needed to update it according to the four different strategies. Thus, for $\Szero$, this will coincide with the time spent for building $\Mzero$ (as, in this case, no further action is taken for updating the model); for $\Sone$, we compute the time spent for building $\Mzero$ plus the time spent for the retraining over $\tr$ (no hyperopt); for $\Stwo$, we compute the time spent for building $\Mzero$ plus the time needed for the retraining over $\tr$ (with hyperopt); and, finally, for $\Sthree$, we compute the time spent for building $\Mzero$ plus the time needed for updating it with the data in $\trone$.

\subsection{Experimental Settings}
The tool used for the experimentation is \nirdi \citep{DBLP:conf/bpm/RizziSFGKM19}.
%
%
The experimental evaluation was performed on a workstation Dell Precision 7820 with the following configuration:
\begin{enumerate*}[label=(\roman*)]
    \item 314GB DDR4 2666MHz RDIMM ECC of RAM;
    \item double Intel Xeon Gold 6136 3.0GHz, 3.7GHz Turbo, 12C, 10.4GT/s 3UPI, 24.75MB Cache, HT (150W) CPU; and
    \item one 2.5" 256GB SATA Class 20 Solid State Drive.
\end{enumerate*}
We assumed to have only 1 CPU for training the predictive models, i.e., we did not parallelise the training of the base learners of the Random Forests.
We also ensured that there was no racing condition over the disk and no starvation over the RAM usage by actively monitoring the resources through Netdata~\citep{netdata}.
Each experiment was allowed to run for at most 100 hours.
No other experiment or interaction with the workstation was performed other than the monitoring of the used resources.


\section{Results}
\label{sec:results}
In this section, we present the results of our experiments, reported in Tables~\ref{tab:accuracy1}--\ref{tab:time1}, and discuss how they allow us to answer the two research questions introduced before. We also provide a discussion about the four update strategies with a cost-effectiveness analysis and an analysis of the validity threats. To ensure reproducibility, the datasets used, the configurations, and the detailed results are available at \url{http://bit.ly/how_do_I_update_my_model}.

\subsection{Discussion}

\paragraph{Answering RQ1.}

The AUC of all models, for the two experimental settings 10\%-70\% and 40\%-40\%, for all datasets, is reported in Tables~\ref{tab:accuracy1_a} and ~\ref{tab:accuracy1_b}. The best result for each dataset and labelling is emphasised in italic\footnote{We indicate in italic the best result, which may be due to digits smaller than the third decimal one reported in the paper.}. Since, in many cases, different models have very close accuracy, we have emphasised in bold the results that differ from the best ones for less than 0.01.
Tables~\ref{tab:accuracy2_a} and~\ref{tab:accuracy2_b} report, for the two experimental settings 10\%-70\% and 40\%-40\%, the percentage of gain/loss of $\Mone$, $\Mtwo$, and $\Mthree$ w.r.t. $\Mzero$. The percentage of gain or loss\footnote{Computed as $\frac{\M_{i}-\Mzero}{\Mzero}$, $i \in \{1,2,3\}$.} is reported together with an histogram of gains and losses. In order to ease the comparison, $\Mtwo$ is reported in both tables.

\begin{table}
	\centering
    \begin{subtable}{.5\linewidth}
		\centering
		\scalebox{.70}{
        \begin{tabular}{@{}c c c c c c@{}}
			\toprule
			\textbf{Dataset} & $ \mathbf{\phi} $ & $\mathbf{\Mzero}$ & $\mathbf{\Mone}$ & $\mathbf{\Mtwo}$ & $\mathbf{\Mthree}$ \\
			\midrule
			\multirow{3}*{BPIC11}   & $\phi _{11}$ & 0.673 & 0.885 & \textit{\textbf{0.919}} & 0.833 \\
								    & $\phi _{12}$ & 0.745 & 0.887 & \textit{\textbf{0.964}} & 0.883 \\
								    & $\phi _{13}$ & 0.843 & \textbf{0.916} & \textbf{0.921} & \textit{\textbf{0.926}} \\
			\cmidrule{2-6}
			\multirow{3}*{BPIC12}   & $\phi _{21}$ & 0.576 & 0.648 & \textit{\textbf{0.702}} & 0.671 \\
								    & $\phi _{22}$ & 0.672 & \textbf{0.733} & \textit{\textbf{0.740}} & 0.676 \\
								    & $\phi _{23}$ & 0.517 & 0.504 & 0.514 & \textit{\textbf{0.561}} \\
			\cmidrule{2-6}
			\multirow{3}*{BPIC15}   & $\phi _{31}$ & 0.908 & 0.916 & 0.935 & \textit{\textbf{0.993}} \\
								    & $\phi _{32}$ & 0.928 & 0.911 & 0.923 & \textit{\textbf{0.972}} \\
								    & $\phi _{33}$ & 0.961 & 0.976 & \textbf{0.988} & \textit{\textbf{0.995}} \\
			\cmidrule{2-6}
			BPIC18                  & $\phi _{41}$ & 0.532 & \textbf{0.999} & \textbf{0.991} & \textit{\textbf{0.999}} \\
			\cmidrule{2-6}

			DriftRIO1 & $\phi _{51}$ & 0.603 & \textbf{0.964} & \textit{\textbf{0.965}} & \textbf{0.964} \\ \cmidrule{2-6}

			DriftRIO2 & $\phi _{61}$ & 0.761 & \textbf{0.856} & \textbf{0.854} & \textit{\textbf{0.857}} \\
			\bottomrule
        \end{tabular}
		}
		\caption{Setting 10\%-70\%.}
		\label{tab:accuracy1_a}
    \end{subtable}%
    \begin{subtable}{.5\linewidth}
		\centering
		\scalebox{.70}{
        \begin{tabular}{@{}c c c c c c@{}}
			\toprule
			\textbf{Dataset} & $ \mathbf{\phi} $ & $\mathbf{\Mzero}$ & $\mathbf{\Mone}$ & $\mathbf{\Mtwo}$ & $\mathbf{\Mthree}$ \\
			\midrule
			\multirow{3}*{BPIC11}   & $\phi _{11}$ & 0.781 & \textit{\textbf{0.935}} & 0.919 & 0.902 \\
								    & $\phi _{12}$ & 0.811 & 0.909 & \textit{\textbf{0.964}} & 0.930 \\
								    & $\phi _{13}$ & 0.894 & \textbf{0.918} & \textit{\textbf{0.921}} & \textbf{0.920} \\
			\cmidrule{2-6}
			\multirow{3}*{BPIC12}   & $\phi _{21}$ & 0.631 & 0.671 & \textit{\textbf{0.702}} & 0.682 \\
								    & $\phi _{22}$ & 0.674 & 0.672 & \textit{\textbf{0.740}} & 0.702 \\
								    & $\phi _{23}$ & 0.509 & 0.511 & 0.514 & \textit{\textbf{0.560}} \\
			\cmidrule{2-6}
			\multirow{3}*{BPIC15}   & $\phi _{31}$ & 0.779 & \textit{\textbf{0.991}} & 0.935 & 0.944 \\
								    & $\phi _{32}$ & 0.895 & 0.907 & 0.923 & \textit{\textbf{0.953}} \\
								    & $\phi _{33}$ & 0.972 & \textit{\textbf{0.994}} & \textbf{0.988} & \textbf{0.987} \\
			\cmidrule{2-6}
			BPIC18                  & $\phi _{41}$ & 0.543 & \textbf{1.000} & \textbf{0.991} & \textit{\textbf{1.000}} \\
			\cmidrule{2-6}
			DriftRIO1 & $\phi _{51}$ & 0.559 & \textbf{0.964} & \textbf{0.965} & \textit{\textbf{0.969}} \\
			DriftRIO2	& $\phi _{61}$ & 0.805 & 0.698 & \textit{\textbf{0.854}} & \textbf{0.841} \\
			\bottomrule
        \end{tabular}
		}
		\caption{Setting 40\%-40\%.}
		\label{tab:accuracy1_b}
	\end{subtable} 
	\caption{The accuracy results.}
	\label{tab:accuracy1}	
\end{table}

\begin{table}
	\centering
    \begin{subtable}{.5\linewidth}
		\centering
		\scalebox{.70}{
        \begin{tabular}{@{}c c c c c@{}} 
			\toprule
			\textbf{Dataset} & $ \mathbf{\phi} $ & $\mathbf{\Mone}$ & $\mathbf{\Mtwo}$ & $\mathbf{\Mthree}$ \\
			\midrule
			\multirow{3}*{BPIC11}   & $\phi _{11}$ & \databar{0.31} & \databar{0.36} & \databar{0.23} \\
								    & $\phi _{12}$ & \databar{0.19} & \databar{0.29} & \databar{0.18} \\
								    & $\phi _{13}$ & \databar{0.08} & \databar{0.09} & \databar{0.09} \\
			\cmidrule{2-5}
			\multirow{3}*{BPIC12}   & $\phi _{21}$ & \databar{0.12} & \databar{0.21} & \databar{0.16} \\
								    & $\phi _{22}$ & \databar{0.09} & \databar{0.10} & \databar{0.00} \\
								    & $\phi _{23}$ & \databarred{-0.02} & \databar{0.00} & \databar{0.08} \\
			\cmidrule{2-5}
			\multirow{3}*{BPIC15}   & $\phi _{31}$ & \databar{0.01} & \databar{0.03} & \databar{0.09} \\
								    & $\phi _{32}$ & \databarred{-0.01} & \databar{0.00} & \databar{0.04} \\
								    & $\phi _{33}$ & \databar{0.01} & \databar{0.02} & \databar{0.03} \\
			\cmidrule{2-5}
			BPIC18                  & $\phi _{41}$ & \databar{0.87} & \databar{0.86} & \databar{0.87} \\
			\cmidrule{2-5}
			DriftRIO1 & $\phi _{51}$ & \databar{0.59} & \databar{0.59} & \databar{0.59} \\
			\cmidrule{2-5}
			DriftRIO2 & $\phi _{61}$ & \databar{0.12} & \databar{0.12} & \databar{0.12} \\
			\bottomrule
        \end{tabular}
		}
		\caption{Setting 10\%-70\%.}
		\label{tab:accuracy2_a}
    \end{subtable}%
    \begin{subtable}{.5\linewidth}
		\centering
		\scalebox{.70}{
        \begin{tabular}{@{}c c c c c@{}}
			\toprule
			\textbf{Dataset} & $ \mathbf{\phi} $ &  $\mathbf{\Mone}$ & $\mathbf{\Mtwo}$ & $\mathbf{\Mthree}$ \\
			\midrule
			\multirow{3}*{BPIC11}   & $\phi _{11}$ & \databar{0.19} & \databar{0.17} & \databar{0.15} \\
								    & $\phi _{12}$ & \databar{0.12} & \databar{0.18} & \databar{0.14} \\
								    & $\phi _{13}$ & \databar{0.02} & \databar{0.03} & \databar{0.03} \\
			\cmidrule{2-5}
			\multirow{3}*{BPIC12}   & $\phi _{21}$ & \databar{0.06} & \databar{0.11} & \databar{0.08} \\
								    & $\phi _{22}$ & \databarred{-0.00} & \databar{0.09} & \databar{0.04} \\
								    & $\phi _{23}$ & \databar{0.00} & \databar{0.01} & \databar{0.10} \\
			\cmidrule{2-5}
			\multirow{3}*{BPIC15}   & $\phi _{31}$ & \databar{0.27} & \databar{0.20} & \databar{0.21} \\
								    & $\phi _{32}$ & \databar{0.01} & \databar{0.03} & \databar{0.06} \\
								    & $\phi _{33}$ & \databar{0.02} & \databar{0.01} & \databar{0.01} \\
			\cmidrule{2-5}
			BPIC18                  & $\phi _{41}$ & \databar{0.84} & \databar{0.82} & \databar{0.84} \\
			\cmidrule{2-5}
			DriftRIO1 & $\phi _{51}$ & \databar{0.72} & \databar{0.72} & \databar{0.73} \\ \cmidrule{2-5}
			DriftRIO2 & $\phi _{61}$ & \databarred{-0.13} & \databar{0.06} & \databar{0.04} \\
			\bottomrule
        \end{tabular}
		}
		\caption{Setting 40\%-40\%.}
		\label{tab:accuracy2_b}
	\end{subtable} 
	\caption{Accuracy improvement against $\Mzero$.}
	\label{tab:accuracy2}	
\end{table}

By looking at the tables, we can immediately see that $\Mtwo$ and $\Mthree$ are the clear best performers, especially in the first setting 10\%-70\%, and that $\Mzero$ is almost consistently the worst performer, often with a significant difference in terms of accuracy. This highlights that the need to update the predictive models is a real issue in typical \ppm settings. The only exception to this finding is provided by the results obtained for the BPIC15 dataset, where the performance of $\Mzero$ is comparable to that of the other models for almost all the outcome formulae. This is due to the fact that, for this dataset, there is a high homogeneity of the process behavior over time. This is confirmed by the entropy values, provided in Table~\ref{table:entropy}, that remain quite stable across the entire log.
Moreover, we can observe that BPIC12 with labelling $\phi_{23}$ has overall the lowest accuracy for all the four update strategies. This is possibly due to the high label unbalance characterizing this dataset (see Table~\ref{table:labels}).

The performance of all the evaluated strategies is overall higher in the 40\%-40\% setting and this is likely related to the higher amount of data used for hyperparameter optimisation. Nonetheless, the lower performance of $\Mzero$ also in the 40\%-40\% setting indicates the need to update the models with new data at regular intervals.
Concerning the possible differences between the results obtained using datasets with and without an explicit Concept Drift, our experiments did not find any striking variation in the different strategies, thus consolidating the finding that devising update strategies is important in general, also in scenarios where the process changes over time are not so definite. Nonetheless, if we look at Tables~\ref{tab:accuracy2_a} and~\ref{tab:accuracy2_b}, it is easy to see that the experiments with an explicit Concept Drift are the ones with the greatest difference between $\Mzero$ and all the other models, thus confirming that an explicit Concept Drift can have a significant negative influence on the performance of a \ppm model, if it is not updated. This is especially true for BPIC18, in which the Concept Drift highly affects the entropy measure (Table~\ref{table:entropy}) in a way that the difference between the entropy value of the split 0\%-40\%-80\%-100\% is higher than the entropy value of the split 40\%-80\%-80\%-100\% for both entropy metrics.

Tables~\ref{tab:accuracy2_a} and~\ref{tab:accuracy2_b} show also another interesting aspect of our evaluation: while $\Mtwo$ and $\Mthree$ tend to always gain against $\Mzero$ (or to be stable in very few cases), the same cannot be said for $\Mone$. In fact, if we look at BPIC12 with labelling $\phi_{23}$ and BPIC15 with labelling $\phi_{32}$ in Table~\ref{tab:accuracy2_a}, and, particularly, at \rio with labelling $\phi_{52}$ in Table~\ref{tab:accuracy2_b}, we can see a decrease of the accuracy. By carrying out a deeper analysis of the chosen hyperparameters, we found that the lower accuracy of $\Mone$ is due to the inappropriateness of the hyperparameters derived from $\trzero$ to the new data used to build $\Mone$. While this aspect may need to be better investigated, we can conclude that, while \textbf{re-train with no hyperopt} is usually a viable solution, it is nonetheless riskier than \textbf{full re-train} or \textbf{incremental update}.

The general findings and trends derived from the results obtained using Random Forest as classifier are further confirmed, with few exceptions, by the results obtained by using Perceptron~\citep{perceptron} as predictive model in the analysis of the four update strategies. The perceptron results are reported in Appendix~\ref{sec:appendix}.

To sum up, concerning \textbf{RQ1}, our evaluation shows that \textbf{full re-train} and \textbf{incremental update} are the best performing update strategies in terms of accuracy, followed by \textbf{re-train with no hyperopt}. With the exception of BPIC15 in the 10\%-70\% setting, \textbf{do nothing} is, often by far, the worst strategy, indicating the importance of updating the predictive models with new data, when it becomes available.


\paragraph{Answering RQ2.}
\label{ssec:RQ3}

The time spent for creating the four models, for the two experimental settings 10\%-70\% and 40\%-40\%, for all datasets, is reported in Tables~\ref{tab:time1_a} and~\ref{tab:time1_b} using the ``hh:mm:ss'' format. The best results for each dataset and labelling (that is the lower execution times) are emphasised in italic, while execution times that differ for less than 60 seconds from the best ones are indicated in bold.
The last two rows of each table report the average time (and standard deviation) necessary to train a model for a given strategy.
In order to ease the comparison, $\Mtwo$ is reported in both tables.
$\Mzero$ and $\Mtwo$ are self-contained models that are ``built from scratch'' and, therefore, the time reported in the tables for these models is the time spent to train them. Differently, $\Mone$ and $\Mthree$ are built in a two-steps fashion that includes a training phase but also the usage of the \hps used to build $\Mzero$. Therefore, their construction time is measured by summing up the time spent for the training phase and the time spent for building $\Mzero$.

\begin{table}
	\centering
    \begin{subtable}{.5\linewidth}
		\centering
		\scalebox{.60}{
     		\begin{tabular}{@{}c c c c c c@{} }
			\toprule
			\textbf{Dataset} & $ \mathbf{\phi} $ & $\mathbf{\Mzero}$ & $\mathbf{\Mone}$ & $\mathbf{\Mtwo}$ & $\mathbf{\Mthree}$ \\
			\midrule
				\multirow{3}*{BPIC11}   & $\phi _{11}$  & \textit{\textbf{05:31:49}} & 05:37:18 & 10:12:42 & 05:35:53 \\
										& $\phi _{12}$  & \textit{\textbf{06:03:05}} & 06:05:01 & 09:23:21 & 06:04:32 \\
										& $\phi _{13}$  & \textit{\textbf{01:00:21}} & \textbf{01:00:25} & 30:43:11 & \textbf{01:00:25} \\
				\cmidrule{2-6}
				\multirow{3}*{BPIC12}   & $\phi _{21}$  & \textit{\textbf{00:46:59}} & 00:51:00 & 08:02:32 & \textbf{00:47:02} \\
										& $\phi _{22}$  & \textit{\textbf{00:46:42}} & \textit{\textbf{00:46:42}} & 09:03:06 & \textbf{00:46:43} \\
										& $\phi _{23}$  & \textit{\textbf{03:38:22}} & \textit{\textbf{03:38:22}} & 11:37:39 & \textbf{03:38:37} \\
				\cmidrule{2-6}
				\multirow{3}*{BPIC15}   & $\phi _{31}$  & \textit{\textbf{00:13:10}} & \textbf{00:13:11} & 00:27:31 & \textbf{00:13:11} \\
										& $\phi _{32}$  & \textit{\textbf{00:14:49}} & \textbf{00:14:52} & 00:51:01 & \textbf{00:14:51} \\
										& $\phi _{33}$  & \textit{\textbf{00:13:03}} & \textbf{00:13:04} & 00:52:12 & \textit{\textbf{00:13:03}} \\
				\cmidrule{2-6}
				BPIC18                  & $\phi _{41}$  & \textit{\textbf{26:44:03}} & 27:26:00 & 74:44:03 & 26:51:24 \\
				\cmidrule{2-6}
				DriftRIO1 & $\phi _{51}$  & \textit{\textbf{00:12:38}} & \textbf{00:12:39} & 00:16:17 & \textbf{00:12:39} \\
				\cmidrule{2-6}
				DriftRIO2	& $\phi _{61}$  & \textit{\textbf{00:13:19}} & \textbf{00:13:21} & 00:14:22 & \textbf{00:13:21} \\
				\specialrule{1pt}{2pt}{2pt}
				\multicolumn{2}{c}{Average Time}   & \textit{\textbf{01:43:07}} & 01:44:10 & 12:52:20 & \textbf{01:43:40} \\
				\multicolumn{2}{c}{Std deviation}  & 02:14:45 & 02:15:53 & 21:06:04 & 02:15:44 \\
				\bottomrule
			\end{tabular}
			}
			\caption{Setting 10\%-70\%.}
			\label{tab:time1_a}
		\end{subtable}%
		\begin{subtable}{.5\linewidth}
			\centering
			\scalebox{.60}{
			\begin{tabular}{c c c c c c }
			\toprule
			\textbf{Dataset} & $ \mathbf{\phi} $ & $\mathbf{\Mzero}$ & $\mathbf{\Mone}$ & $\mathbf{\Mtwo}$ & $\mathbf{\Mthree}$ \\
			\midrule
				\multirow{3}*{BPIC11}   & $\phi _{11}$  & \textit{\textbf{05:00:33}} & \textbf{05:00:38} & 10:12:42 & \textbf{05:00:36} \\
										& $\phi _{12}$  & 18:53:46 & 18:54:40 & \textit{\textbf{09:23:21}} & 18:54:14 \\
										& $\phi _{13}$  & \textit{\textbf{05:13:21}} & \textbf{05:13:38} & 30:43:11 & \textbf{05:13:29} \\
				\cmidrule{2-6}
				\multirow{3}*{BPIC12}   & $\phi _{21}$  & \textit{\textbf{03:23:40}} & 03:27:06 & 08:02:32 & \textbf{03:23:42} \\
										& $\phi _{22}$  & \textit{\textbf{03:40:02}} & \textit{\textbf{03:40:02}} & 09:03:06 & \textbf{03:40:03} \\
										& $\phi _{23}$  & \textit{\textbf{05:00:07}} & \textbf{05:00:08} & 11:37:39 & \textbf{05:00:25} \\
				\cmidrule{2-6}
				\multirow{3}*{BPIC15}   & $\phi _{31}$  & 00:29:11 & 00:29:16 & \textit{\textbf{00:27:31}} & 00:29:15 \\
										& $\phi _{32}$  & \textit{\textbf{00:31:51}} & \textbf{00:31:54} & 00:51:01 & \textbf{00:31:53} \\
										& $\phi _{33}$  & \textit{\textbf{00:35:26}} & \textbf{00:35:27} & 00:52:12 & \textit{\textbf{00:35:26}} \\
				\cmidrule{2-6}
				BPIC18                  & $\phi _{41}$  & \textit{\textbf{55:06:43}} & 55:08:42 & 74:44:03 & \textbf{55:06:57} \\
				\cmidrule{2-6}
				DriftRIO1 & $\phi _{51}$  & \textit{\textbf{00:14:39}} & \textbf{00:14:41} & 00:16:17 & \textbf{00:14:41} \\
				\cmidrule{2-6}
				DriftRIO2	& $\phi _{61}$  & \textit{\textbf{00:13:53}} & \textit{\textbf{00:13:53}} & \textbf{00:14:22} & \textit{\textbf{00:13:53}} \\
				\specialrule{1pt}{2pt}{2pt}
				\multicolumn{2}{c}{Average Time} & \textit{\textbf{03:56:03}} & \textbf{03:56:29} & 12:52:20 & \textbf{03:56:09} \\
				\multicolumn{2}{c}{Std deviation}& 05:23:12 & 05:23:24 & 21:06:04 & 05:23:20 \\
				\bottomrule
			\end{tabular}
			}
		\caption{Setting 40\%-40\%.}
		\label{tab:time1_b}
	\end{subtable} 
	\caption{The time results.}
	\label{tab:time1}	
\end{table}

By looking at Tables~\ref{tab:time1_a} and~\ref{tab:time1_b}, we can immediately see that, among all the evaluated strategies, $\Mzero$ is the clear best performer, especially for the 10\%-70\% setting, and that $\Mtwo$ is almost consistently the worst performer, often with a significant difference in terms of time spent to build the model (with few exceptions in the 40\%-40\% case that we will discuss below).
As a second general observation, we note that $\Mone$ and $\Mthree$ share almost the same construction time with $\Mzero$. This fact is not particularly surprising, as the hyperparameter optimisation routine is often the most expensive step in the construction of this type of predictive models. Therefore, the two strategies $\Sone$ and $\Sthree$ that underline the construction of these models are highly inexpensive, when we consider the time dimension, especially when $\Mzero$ is already available.

If we compare the two experimental settings 10\%-70\% and 40\%-40\%, we can observe that, while $\Mzero$ is almost always the best performer in both settings, the difference between the construction time of $\Mzero$ (and thus of $\Mone$ and $\Mthree$) and the construction time of $\Mtwo$ is significantly higher for the 10\%-70\% setting. While the investigation of \emph{when} it is convenient to perform a \textbf{full re-train} is out of the scope of the paper and is left to further investigations, this finding emphasises the fact that the cost of a \textbf{full re-train} may increase in a significant manner if the update of the predictive model over-delays and the amount of new data greatly increases.
Interestingly enough, the 40\%-40\% setting presents two cases in which $\Mtwo$ is the fastest model to be built. The case of BPIC11 with labelling $\phi_{12}$ is likely due to an ``unfortunate'' guess-estimate in the hyperparameter optimisation step for $\Mzero$, which makes the training time explode\footnote{Note that an explosion of the training time for the same reason affects also model $\Mtwo$ for BPIC11 with labelling $\phi_{13}$.}; the case of BPIC15 with labelling $\phi_{31}$, instead, represents a situation in which $\Mzero$ and $\Mtwo$ take almost the same time to be built.
Concerning possible differences between datasets with and without an explicit Concept Drift, our experiments did not find any striking difference among the evaluated strategies.

Finally, our evaluation did not find any fixed correlation between the training times for all strategies and (i) the size of the \ds and the alphabet of the \ds within the same settings, or (ii) the quality of the predictive model in terms of accuracy (and thus the difficulty of the prediction problem). As an example of the first, we can observe that BPIC12 contains four times the cases of BPIC11, nonetheless, most of the prediction models built for BPIC11 take more time to be constructed than the ones built for BPIC12. Similarly, BPIC15 has an alphabet with a number of activities that is almost 10 times the one of BPIC2018 but the prediction models built for BPIC18 take much more time to be constructed than the ones built for BPIC15. As an example of the second, we can observe, from Table \ref{tab:time1_a}, that the time needed for building $\Mzero$ for BPIC11 with labelling $\phi_{23}$ is much greater than the one needed for building $\Mzero$ for BPIC11 with labelling $\phi_{13}$, even if the accuracy for the same cases, in Table~\ref{tab:accuracy1_a}, follows the inverse trend.\footnote{This is just one of many samples of an inverse relation between time and accuracy that can be found in the tables.}


To sum up, concerning \textbf{RQ2}, our evaluation shows that, once $\Mzero$ is available, \textbf{incremental update} and \textbf{re-train with no hyperopt} are the two most convenient update strategies - as they can be built in almost no time. This may suggest the possibility to implement an almost continuous update strategy whenever new data becomes available.
While the investigation of \emph{when} it is convenient to perform a \textbf{full re-train} is out of the scope of the paper, our experiments show that the cost of a \textbf{full re-train} may increase in a significant manner if the update of the predictive model over-delays and the amount of new data increases significantly.

\begin{figure}
	\centering
\begin{subfigure}{.48\textwidth}
  \centering
  \includegraphics[width=\linewidth]{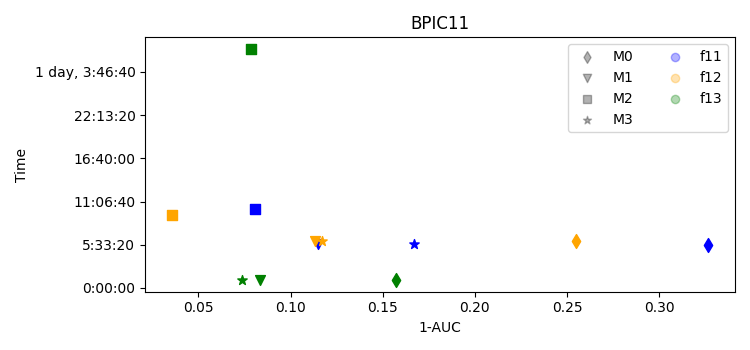}
  \caption{BPIC11 10\%.}
  \label{fig:pp_bpic11_10}
\end{subfigure}
\begin{subfigure}{.48\textwidth}
  \centering
  \includegraphics[width=\linewidth]{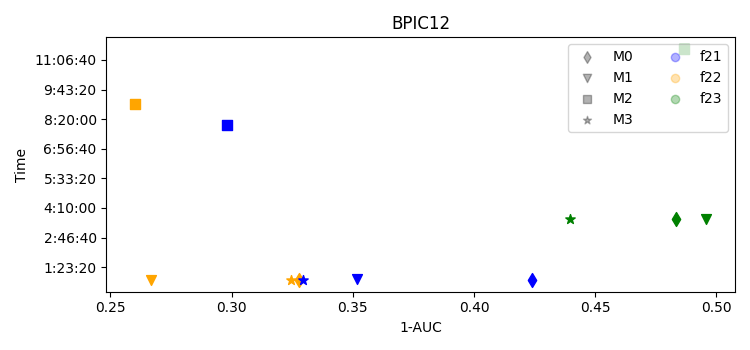}
  \caption{BPIC12 10\%.}
  \label{fig:pp_bpic12_10}
\end{subfigure}\\ %
\begin{subfigure}{.48\textwidth}
  \centering
  \includegraphics[width=\linewidth]{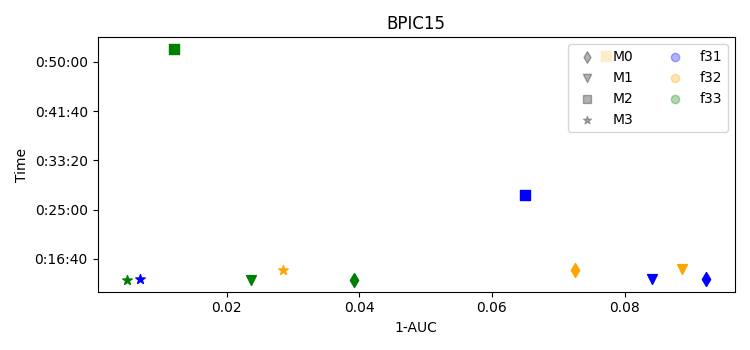}
  \caption{BPIC15 10\%.}
  \label{fig:pp_bpic15_10}
\end{subfigure}%
\begin{subfigure}{.48\textwidth}
  \centering
  \includegraphics[width=\linewidth]{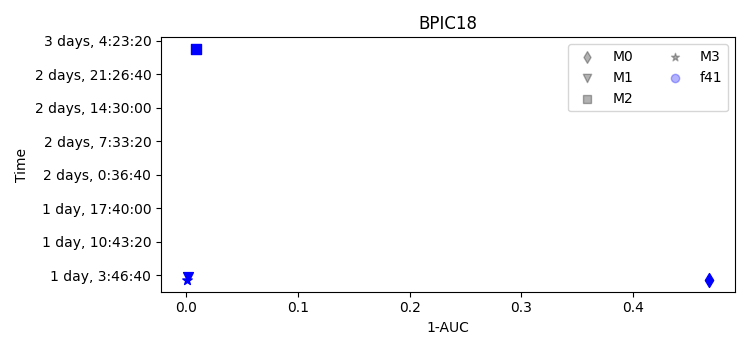}
  \caption{BPIC18 10\%.}
  \label{fig:pp_bpic18_10}
\end{subfigure} \\%
\begin{subfigure}{.48\textwidth}
  \centering
  \includegraphics[width=\linewidth]{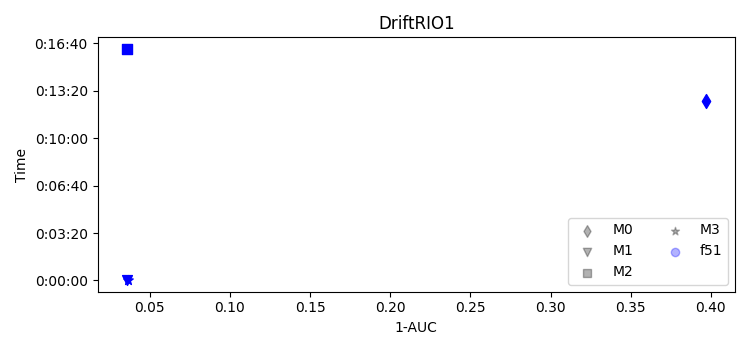}
  \caption{\rio1 10\%.}
  \label{fig:pp_rio1_10}
\end{subfigure}%
\begin{subfigure}{.48\textwidth}
  \centering
  \includegraphics[width=\linewidth]{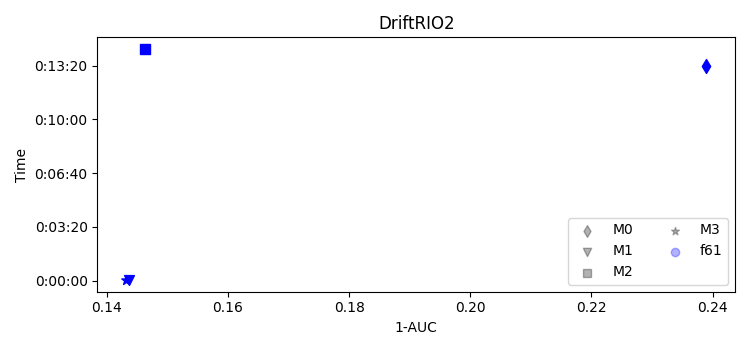}
  \caption{\rio2 10\%.}
  \label{fig:pp_rio2_10}
\end{subfigure}%
	\caption{Inaccuracy versus time plots (10\%).}
	\label{fig:pareto_plots_10}
\end{figure}

\begin{figure}
	\centering
\begin{subfigure}{.48\textwidth}
  \centering
  \includegraphics[width=\linewidth]{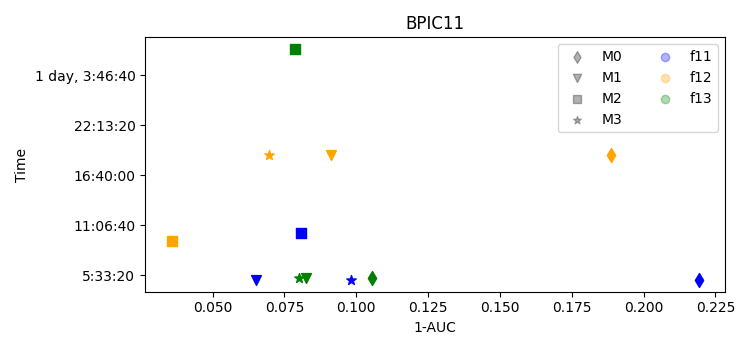}
  \caption{BPIC11 40\%.}
  \label{fig:pp_bpic11}
\end{subfigure}
\begin{subfigure}{.48\textwidth}
  \centering
  \includegraphics[width=\linewidth]{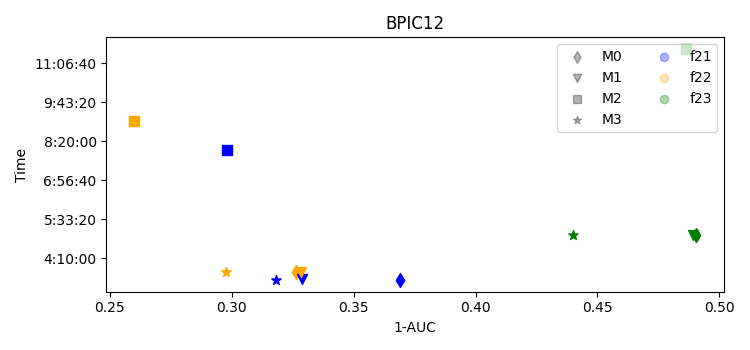}
  \caption{BPIC12 40\%.}
  \label{fig:pp_bpic12}
\end{subfigure}\\ %
\begin{subfigure}{.48\textwidth}
  \centering
  \includegraphics[width=\linewidth]{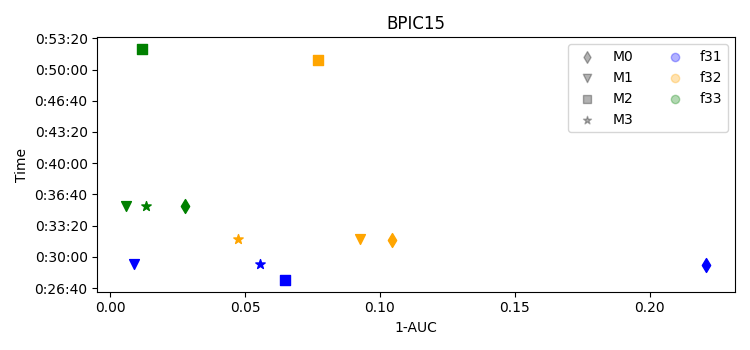}
  \caption{BPIC15 40\%.}
  \label{fig:pp_bpic15}
\end{subfigure}%
\begin{subfigure}{.48\textwidth}
  \centering
  \includegraphics[width=\linewidth]{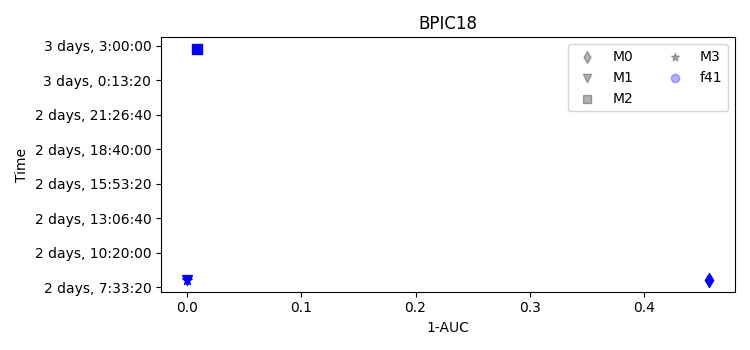}
  \caption{BPIC18 40\%.}
  \label{fig:pp_bpic18}
\end{subfigure} \\%
\begin{subfigure}{.48\textwidth}
  \centering
  \includegraphics[width=\linewidth]{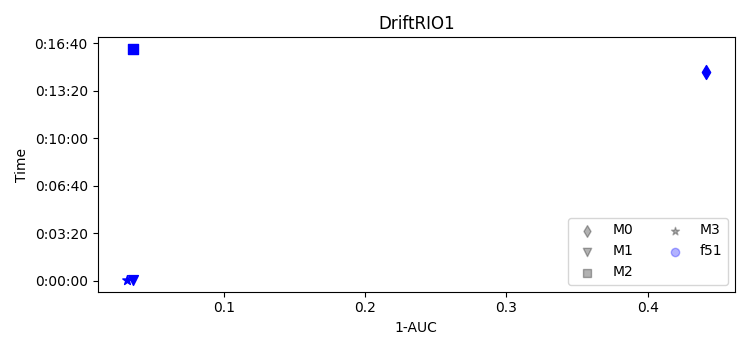}
  \caption{\rio1 40\%.}
  \label{fig:pp_rio1}
\end{subfigure}%
\begin{subfigure}{.48\textwidth}
  \centering
  \includegraphics[width=\linewidth]{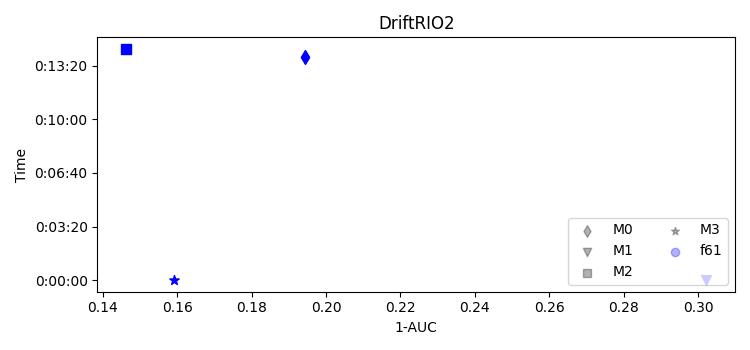}
  \caption{\rio2 40\%.}
  \label{fig:pp_rio2}
\end{subfigure}%
	\caption{Inaccuracy versus time plots (40\%).}
	\label{fig:pareto_plots}
\end{figure}

\paragraph{Overall Conclusions.}
The plots in Figures~\ref{fig:pareto_plots_10} and~\ref{fig:pareto_plots} show inaccuracy and time related to the 10\%-70\% and 40\%-40\% settings, respectively, for each of the considered datasets. The closer the item is to the origin, the best is the balance between the time required for training, re-training, or updating the model and the accuracy of the results. By looking at the plots, it is clear that the worst choice in terms of balance is given by $\Mzero$, while, for the other three models, the choice somehow depends on the dataset and on the labelling. With the only exception of $\phi_{12}$, $\phi_{21}$, $\phi_{22}$, and $\phi_{33}$ for both settings, as well as of $\phi_{11}$ for the 10\%-70\% setting and of $\phi_{61}$ for the 40\%-40\% setting, for all other datasets and labellings, $\Mone$ and/or $\Mthree$ are the only non-dominated update strategies, i.e., those strategies for which another strategy improving both the inaccuracy and the time dimension does not exist.\footnote{We did not take into account small differences that cannot be observed with the naked eye in the plots.}

To conclude, our evaluation shows that the \textbf{do-nothing} strategy is not a viable strategy as the accuracy performance of a non-updated model tends to significantly decrease for typical real-life datasets (with and without explicit Concept Drift), whereas lightweight update strategies, such as the \textbf{incremental update} and \textbf{re-train with no hyperopt}, are, instead, often extremely effective in updating the models.
\textbf{Full re-train} offers a strategy that almost always achieves the best accuracy (or an accuracy in line with the best one). Nonetheless, its training time may increase significantly, especially in the presence of an abundance of new data. According to our experiments, the \textbf{incremental update} is able to keep up with the \textbf{full re-train} strategy and deliver a properly fitted model almost in real time, suggesting that the potential of incremental models is under-appreciated in \ppm, and smart \ppm solutions could be developed leveraging this update strategy.

\subsection{Cost-effectiveness analysis}
In order to have a better grasp of the cost-effectiveness of the different update strategies, we also investigated the costs required by the update strategies, when new batches of data become available along the time.
In particular, given a set of batches of train sets $\trzero$, $\trone$, \ldots ${\cal TR}_n$, and the corresponding batches of test sets $\te_{0}$, $\te_{1}$, \ldots $\te_n$ (where $\te_{0}$ is the test set immediately following $\trzero$, $\te_{1}$ the test set immediately following $\trone$ in the temporal timeline), we can define $CE_M$($\trzero$, $\trone$, \ldots ${\cal TR}_n$, $\te_{0}$, $\te_{1}$, \ldots $\te_n$) -- from here on shortened as $CE_M$(${\cal TR}_{0 \ldots n}$, $\te_{0 \ldots n}$)-- as the cost of a model trained with $n$ batches of arriving data $\trzero$, $\trone$, \ldots ${\cal TR}_n$, and tested with the corresponding batches of test data $\te_{1}$, \ldots $\te_n$.

In our scenario, the cost-effectiveness of the update strategies is characterized by two main aspects: on the one hand, the cost of the time required for building the model and, on the other, the cost of returning wrong predictions (prediction inaccuracy).
We can hence define $CE_{\cal M}$(${\cal TR}_{0 \ldots n}$, $\te_{0 \ldots n}$) as the sum of (i) $CT_{\cal M}$(${\cal TR}_{0 \ldots n}$, $\te_{0 \ldots n}$), i.e., the cost of the time required for building, training and, when necessary, retraining the model $M$, whenever a new batch of traces arrives;\footnote{We assume that the time cost for providing predictions at runtime, i.e., for testing our model is negligible.} and of (ii) $CI_{\cal M}$(${\cal TR}_{0 \ldots n}$, $\te_{0 \ldots n}$), i.e., the cost of the inaccuracy due to wrong predictions returned by the trained models on the traces of the test sets. Low values for a given model indicate a good cost-effectiveness model.

Defining $CT_{T}(TR)$, $CT_{T}^{H}(TR)$, and $CT_{U}(TR)$ as the time required for training, training and optimizing the hyperparameters, and incrementally updating a model with the train set $TR$, respectively, the time costs related to the four models can be computed as reported in Equation~\ref{eq:CT}. The time cost for $\Mzero$ is given by the only cost required for training and optimizing the hyperparameters on $\trzero$. For $\Mone$ ($\Mthree$), besides the cost for training and optimizing the hyperparameters on $\trzero$, when the i-th train set batch is available, the costs for training (updating) the model with the union of the train set batches up to the i-th one (with the i-th train batch), have also to be considered. Finally, the time cost for $\Mtwo$ is given by the cost required for re-training and optimizing the hyperparameters from scratch each time a new batch arrives.


\begin{small}
\begin {equation} \label {eq:CT}
\begin{split}
CT_{\Mzero}({\cal TR}_{0 \ldots n}, \te_{0 \ldots n})&=CT_{T}^{H}(\trzero)\\
CT_{\Mone}({\cal TR}_{0 \ldots n}, \te_{0 \ldots n})&=CT_{T}^{H}(\trzero)+\sum_{i=1}^{n-1} CT_T(\bigcup_{j=1}^{i}{\cal TR}_i) \\
CT_{\Mtwo}({\cal TR}_{0 \ldots n}, \te_{0 \ldots n})&=\sum_{i=0}^{n-1} CT_T^{H}(\bigcup_{j=0}^{i}{\cal TR}_i)\\
CT_{\Mthree}({\cal TR}_{0 \ldots n}, \te_{0 \ldots n})&=CT_{T}^{H}(\trzero)+\sum_{i=1}^{n-1} CT_U({\cal TR}_i)
\end{split}
\end{equation}
\end{small}

The inaccuracy costs are instead reported in Equation~\ref{eq:CI}. For $\Mzero$, the inaccuracy cost is given by the sum of the costs obtained by providing predictions using the model trained on the batch $\Mzero$ and tested on each new test set batch $\te_{i}$. For the other three models, instead, the cost of inaccuracy, at the arrival of the train set batch ${\cal TR}_{i}$, is given by the inaccuracy cost obtained from models trained, updated and/or optimized on a train set that takes into account the information on the data of all train set batches up to ${\cal TR}_{i}$, and evaluated on the test set batch $\te_{i}$. The inaccuracy cost is given by the sum of the costs for each new training and test batch ${\cal \tr}_i$ and ${\te}_i$.

\begin{small}
\begin {equation} \label {eq:CI}
\begin{split}
CI_{\Mzero}({\cal TR}_{0 \ldots n}, \te_{0 \ldots n})
&=\sum_{i=0}^{n-1} CI_{\Mzero}({\cal TR}_0,\te_{i})\\
CI_{\Mone}({\cal TR}_{0 \ldots n}, \te_{0 \ldots n})
&=\sum_{i=0}^{n-1} CI_{\Mone}(\bigcup_{j=0}^{i}{\cal TR}_{j}, \te_{i})\\
CI_{\Mtwo}({\cal TR}_{0 \ldots n}, \te_{0 \ldots n})
&=\sum_{i=0}^{n-1} CI_{\Mtwo}(\bigcup_{j=0}^{i}{\cal TR}_{j}, \te_{i})\\
CI_{\Mthree}({\cal TR}_{0 \ldots n}, \te_{0 \ldots n})
&=\sum_{i=0}^{n-1} CI_{\Mthree}(\bigcup_{j=0}^{i}{\cal TR}_{j}, \te_{i})\\
\end{split}
\end{equation}
\end{small}

We assume we can approximate the inaccuracy cost of the model $\Mzero$ tested on the i-th test set batch $\te_i$ -- $CI_{\Mzero}({\cal TR}_0,\te_{i})$ -- with the inaccuracy cost of $\Mzero$ tested on $\te_1$ plus an extra inaccuracy cost $\delta_i^{\Mzero}$, i.e., $CI_{\Mzero}({\cal TR}_0,\te_{i})=CI_{\Mzero}$(${\cal TR}_0,\te_{1})+\delta_i^{\Mzero}$. Similarly, the inaccuracy costs of the other three models -- $CI_{\cal M}(\bigcup_{j=0}^{i}{\cal TR}_{j}, \te_{i})$ -- can be approximated with the inaccuracy cost computed on the first test set plus an extra inaccuracy cost $\delta_i^{\cal M}$, i.e., $CI_{\cal M}$($\bigcup_{j=0}^{i}{\cal TR}_{j}$,\\$ \te_{i})$=$CI_{\cal M}$($\bigcup_{j=0}^{i}{\cal TR}_{j}$,$\te_{1}$)$+\delta_i^{\cal M}$.


Defining $c_t$ as the time hourly cost and $c_e$ as the unary prediction error cost, we can compute the time cost and the inaccuracy cost of a model ${\cal M}$ starting from the time required for training and updating the model $T_{\cal M}$ and from the number of prediction errors (i.e., false positive + false negatives) $E_{\cal M}$ as:

\setlength{\abovedisplayskip}{1pt}
\begin {equation} \label {eq:C3}
\begin{split}
CT_{\cal M}({\cal TR}, \te)&=c_t * T_{\cal M}\\
CI_{\cal M}({\cal TR}, \te)&=c_e * E_{\cal M}\\
\end{split}
\end{equation}

In order to have an estimate of the costs in our scenario, we instantiated such a cost-effectiveness framework. In detail, we set the extra-inaccuracy costs to $0$ ($\delta_i^{\cal M}=0$) and chose a couple of sample configurations for the prediction error unitary costs and for the hourly unitary cost ($c_t=0.1$, $c_e=100$ and $c_t=100$, $c_e=0.1$), as well as for the number of batches $n$ ($n=1$ and $n=5$). Table~\ref{tab:table_costs} reports the obtained results using as reference values the ones of the experimental setting 40\%-40\%. The best result among the four strategies for each outcome is emphasised in italic, while the results that differ from the best ones for less than 1 are emphasised in bold.

The results in the table show that, depending on the unitary costs of time and prediction errors, differences can exist in the choice of the cheapest model. In both settings, $\Mthree$ seems to be the cheapest model for most of the labellings and for different numbers of batches. In the setting in which $c_t=0.1$ and $c_e=100$, due to the low hourly unitary cost, $\Mtwo$ is the cheapest model for some of the labellings. Moreover, no significant differences exist in terms of costs when more and more data batches arrive (at least for the specific assumptions made for this cost-effectiveness framework). In the other setting, i.e., when the time cost is much higher than the error cost ($c_t=100$ and $c_e=0.1$), $\Stwo$ is always the most expensive update strategy, due to its substantial training time. Moreover, in this setting, the cheapest update strategy can change when the number of arriving data batches increases. Indeed, while with only one batch of data, for some of the labellings (e.g., $\phi_{12}$), $\Mone$ is slightly cheaper than or has the same cost as $\Mthree$, in the long run, $\Mthree$ is cheaper than $\Mone$.

To sum up, this instantiation of the cost-effectiveness framework confirms the results of the plots reported in Figure~\ref{fig:pareto_plots}, i.e., that overall $\Sone$ and $\Sthree$ are the strategies providing the best balance between time and accuracy and hence the cheapest update strategies. Moreover, the analysis suggests that, based on the unitary costs of time and errors, as well as on the number of available data batches, differences can exist related to the best update strategy, although $\Mthree$ seems to be consistently cheaper for most of the tested settings.

\begin{table}
	\centering
    \begin{subtable}{0.49\linewidth}
		\centering
		\scalebox{.6}{
      \begin{tabular}{@{}c c c c c c @{}}
			\toprule
			\textbf{Dataset}& $ \mathbf{\phi} $ & $\mathbf{\Mzero}$ & $\mathbf{\Mone}$ & $\mathbf{\Mtwo}$ & $\mathbf{\Mthree}$ \\
			\midrule
			\multirow{3}*{BPIC11}	& $\phi_{11}$ & 7800.5 	 & \textit{\textbf{4600.5}} & 4601.52 									& 4800.5  \\
														& $\phi_{12}$ & 5001.89  &	3401.89 								&	\textit{\textbf{2502.83}} & 4301.89 \\
														& $\phi_{13}$ & 4300.52  & 3900.52 									& \textit{\textbf{3501.19}} & 4100.52 \\
			\cmidrule{2-6}
			\multirow{3}*{BPIC12} & $\phi_{21}$ & 46600.34 & 36500.35									& 31001.14 									& \textit{\textbf{10900.34}} \\
														& $\phi_{22}$ & 14800.37 & 13400.37 								& 6601.27 									 & \textit{\textbf{5400.37}} \\
														& $\phi_{23}$ & 20900.5  & 274000.5 								& \textit{\textbf{19701.66}}& 20900.5 \\
			\cmidrule{2-6}
			\multirow{3}*{BPIC15} & $\phi_{31}$ & 5400.05  & 6000.5 									& 6200.1 										& \textit{\textbf{2900.05}} \\
														& $\phi_{32}$ & 3800.05  & 4100.05 									& 3600.14 									 & \textit{\textbf{3400.05}} \\
														& $\phi_{33}$ & 2900.06  & 3300.06 									& 3100.15 									 & \textit{\textbf{300.06}} 	\\
			\cmidrule{2-6}
			BPIC18        				& $\phi_{41}$ & 236200.8 & 2600.81 									& 20901.08 									&	 \textit{\textbf{300.81}} \\
			\cmidrule{2-6}
			DriftRIO1 						& $\phi_{51}$ & 58200.02 & \textit{\textbf{11000.02}}& 11400.05 								&	\textit{\textbf{11000.02}} \\
			\cmidrule{2-6}
			DriftRIO2 						& $\phi_{61}$ &\textit{\textbf{7900.02}} &	\textit{\textbf{7900.02}} &	\textbf{7900.05} &	\textit{\textbf{7900.02}}  \\
			\bottomrule
     \end{tabular}
		}
		\caption{n=1, $c_e=100$, and $c_t=0.1$.}
		\label{tab:table_costs_a}
    \end{subtable}
    \begin{subtable}{0.49\linewidth}
		\centering
		\scalebox{.6}{
      \begin{tabular}{@{}c c c c c c@{}}
			\toprule
			\textbf{Dataset} & $\mathbf{\phi}$ & $\mathbf{\Mzero}$ & $\mathbf{\Mone}$ & $\mathbf{\Mtwo}$ & $\mathbf{\Mthree}$ \\
			\midrule
			\multirow{3}*{BPIC11}   & $\phi _{11}$ & 39500.5 	& \textit{\textbf{23000.5}} 						&	23015.82 									 & 24000.5\\
															& $\phi _{12}$ & 25001.89 & 17001.91 															 &	\textit{\textbf{12516}} 	& 21501.9 \\
															& $\phi _{13}$ & 21500.52 &	19500.53 															 &	\textit{\textbf{17510.6}} & 20500.52 \\
			\cmidrule{2-6}
			\multirow{3}*{BPIC12}   & $\phi _{21}$ & 233000.3 &	182500.4 & 155012.4 									& \textit{\textbf{54500.34}} \\
															& $\phi _{22}$ & 74000.37 &	67000.37 & 33024.94 									& \textit{\textbf{27000.37}} \\
															& $\phi _{23}$ & 104500.5 & 137000.5 & \textit{\textbf{98517.94}} & 104500.5 \\
			\cmidrule{2-6}
			\multirow{3}*{BPIC15} 	& $\phi _{31}$ & 27000.05 &	30000.05 & 30500.74 									&	\textit{\textbf{14500.05}} \\
															& $\phi _{32}$ & 19000.05 &	20500.05 & 18001.33 									&	 \textit{\textbf{17000.05}} \\
															& $\phi _{33}$ & 14500.06 & 16500.06 & 15501.36 									& \textit{\textbf{1500.06}}\\
			\cmidrule{2-6}
			BPIC18        					& $\phi _{41}$ & 1181001 	& 13000.86 &	57000.43 									&	\textit{\textbf{1500.813}} \\
			\cmidrule{2-6}			
			DriftRIO1 							& $\phi _{51}$ & 291000 	& 55000.03 &	57000.43 									& \textit{\textbf{55000.02}} \\
			\cmidrule{2-6}
			DriftRIO2 							& $\phi _{61}$ & \textit{\textbf{39500.02}} & \textit{\textbf{39500.02}} &	\textbf{39500.38} & \textit{\textbf{39500.02}} \\
			\bottomrule
        \end{tabular}
		}
		\caption{n=5, $c_e=100$, and $c_t=0.1$.}
		\label{tab:table_costs_b}
    \end{subtable}
\\%
    \begin{subtable}{0.49\linewidth}
		\centering
		\scalebox{.6}{
       \begin{tabular}{@{}c c c c c c@{}}
			\toprule
			\textbf{Dataset} & $ \mathbf{\phi} $ & $\mathbf{\Mzero}$ & $\mathbf{\Mone}$ & $\mathbf{\Mtwo}$ & $\mathbf{\Mthree}$ \\
			\midrule
			\multirow{3}*{BPIC11} & $\phi _{11}$ & 508.72 									&	\textit{\textbf{505.66}} 	& 1526.68 & \textbf{505.8} 		 \\
														& $\phi _{12}$ & 1894.61 									& \textit{\textbf{1894.51}} & 2831.03 & \textbf{1894.69}	\\
														& $\phi _{13}$ & \textit{\textbf{526.55}} & \textbf{526.65}						& 1197.72 &	 \textbf{526.6} 		\\
			\cmidrule{2-6}
			\multirow{3}*{BPIC12} & $\phi _{21}$ & 386.04 									&	381.67 										& 1174.67 & \textit{\textbf{350.4}} \\
														& $\phi _{22}$ & 381.52 									&	380.12 										 & 1278.49 & \textit{\textbf{372.15}}\\
														& $\phi _{23}$ & \textit{\textbf{521.09}} &	527.59 										& 1682.64 & \textbf{521.59}	 \\
			\cmidrule{2-6}
			\multirow{3}*{BPIC15} & $\phi _{31}$ & 54.04 										& 54.78 										& 100.6 	 & \textit{\textbf{51.62}}\\
														& $\phi _{32}$ & \textbf{56.88} 					& \textbf{57.27} 						& 141.71 	&	\textit{\textbf{56.54}}\\
														& $\phi _{33}$ & 61.96 										& 62.38 										 & 149.16 	&	\textit{\textbf{59.38}}\\
			\cmidrule{2-6}
			BPIC18        				& $\phi _{41}$ & 1047.42 									& 817.13 										& 1105.54 &	\textit{\textbf{811.91}}\\
			\cmidrule{2-6}
			DriftRIO1 						& $\phi _{51}$ & 82.62 										& \textit{\textbf{35.47}} 	& 62.96 	& \textit{\textbf{35.47}}\\
		\cmidrule{2-6}
			DriftRIO2 						& $\phi _{61}$ &\textit{\textbf{31.04}} 	&\textit{\textbf{31.04}} 		&	54.98 	&	\textit{\textbf{31.04}}\\
			\bottomrule
        \end{tabular}
		}
		\caption{n=1, $c_e=0.1$, and $c_t=100$.}
		\label{tab:table_costs_c}
	\end{subtable}
  \begin{subtable}{0.49\linewidth}
		\centering
		\scalebox{.6}{
       \begin{tabular}{@{}c c c c c c@{}}
			\toprule
			\textbf{Dataset}& $ \mathbf{\phi} $ & $\mathbf{\Mzero}$ & $\mathbf{\Mone}$ & $\mathbf{\Mtwo}$ & $\mathbf{\Mthree}$ \\
			\midrule
			\multirow{3}*{BPIC11} & $\phi _{11}$ & 539.92 										&	\textbf{526} 	&	15841.42 & \textit{\textbf{525.33}} \\
														& $\phi _{12}$ & \textit{\textbf{1914.61}} 	& 1929.11 			& 15985.86 &	\textbf{1915} \\
														& $\phi _{13}$ & \textit{\textbf{543.75}} 	& 549.25 				&	10619.33 & 544 \\
			\cmidrule{2-6}
			\multirow{3}*{BPIC12} & $\phi _{21}$ & 572.44 										& 607.78				& 12557.78 & \textit{\textbf{394.22}} \\
														& $\phi _{22}$ & 440.72 										& 433.72 				&	 13977.22 & \textit{\textbf{393.86}} \\
														& $\phi _{23}$ & \textit{\textbf{604.69}} 	& 637.19 				&	18039.94 & 607.19 \\
			\cmidrule{2-6}
			\multirow{3}*{BPIC15} & $\phi _{31}$ & 75.64 											& 80.72 				& 767.06 		&	\textit{\textbf{63.56}} \\
														& $\phi _{32}$ & 72.08											& 74.83 				&	1346.5 		 & \textit{\textbf{70.36}} \\
														& $\phi _{33}$ & 73.56											& 75.97 				&	1379.56 	 &	\textit{\textbf{60.69}}\\
			\cmidrule{2-6}
			BPIC18        				& $\phi _{41}$ & 1992.22 										& 873.81 				&	5016.97 	&	 \textit{\textbf{814.67}} \\
			\cmidrule{2-6}
			DriftRIO1 						& $\phi _{51}$ & 315.42 										&	\textbf{80.25} & 488.5		& \textit{\textbf{79.69}} \\
		\cmidrule{2-6}
			DriftRIO2 						& $\phi _{61}$ & \textit{\textbf{62.64}}   &	\textit{\textbf{62.64}} &	421.81 & \textit{\textbf{62.64}} \\
			\bottomrule
        \end{tabular}
		}
		\caption{n=5, $c_e=0.1$, and $c_t=100$.}
		\label{tab:table_costs_d}
	\end{subtable}
	\caption{Cost-effectiveness framework instantiation with $\delta_i^{\cal M}=0$ and experimental setting 40\%-40\%.}
	\label{tab:table_costs}	
\end{table}

\subsection{Threats to Validity}

The main threats affecting the validity of the evaluation carried out are external validity threats, limiting the generalizability of the results. 
Indeed, although we investigated the usage of different update strategies on different types of labellings, we limited the investigation to outcome predictions and to classification techniques typically used with this type of predictions. We plan to inspect other types of predictions, i.e., numeric and sequence predictions, together with typical techniques used with them, i.e., regression and deep learning techniques, for future work.

Finally, the lack of an exhaustive investigation of the \hp values affects the construction validity of our experimentation. We limited this threat by using standard techniques for \hp optimisation~\citep{DBLP:conf/icml/BergstraYC13}.

\section{Related Work}
\label{sec:related}
To the best of our knowledge, no other work exists on the comparison of update strategies for Predictive Process Monitoring models with the exception of the two by \citet{DBLP:conf/bpm/PauwelsC21} and \citet{DBLP:conf/IEEEscc/MaisenbacherW17}. We hence first position our work within the \ppm field and then address a specific comparison with \citet{DBLP:conf/bpm/PauwelsC21} and \citet{DBLP:conf/IEEEscc/MaisenbacherW17}.

We can classify \ppm works based on the types of predictions they provide.
A first group of approaches deals with numeric predictions, and, in particular, predictions related to time~\citep{DBLP:journals/is/AalstSS11, DBLP:conf/otm/FolinoGP12, DBLP:conf/icsoc/Rogge-SoltiW13}. A second group of approaches focuses on the prediction of next activities. These approaches mainly use deep learning techniques -- specifically techniques based on LSTM neural networks~\citep{Tax2017, DBLP:conf/bpm/Francescomarino17, DBLP:conf/bpm/0001DR19, brunk2020exploring, DBLP:conf/bpm/TaymouriREBV20}. These studies have shown that when the \dss are large, deep learning techniques can outperform techniques based on classical Machine Learning techniques.
A third group of approaches deals with outcome predictions~\citep{DBLP:journals/tkdd/TeinemaaDRM19, DBLP:conf/caise/MaggiFDG14, DBLP:journals/tsc/Francescomarino19, DBLP:conf/bpm/LeontjevaCFDM15}, which are the ones we focus on.
%
A key difference between these works and the work presented in this paper is that we do not aim at proposing/supporting a specific outcome prediction method, rather we aim at evaluating different update strategies.


The work by~\citet{DBLP:conf/bpm/PauwelsC21} leverages deep learning models to address the challenge of next activity prediction in the context of incremental \ppm. The goal of their paper is two-fold: they explore different strategies to update a model over time for next-activity prediction and they investigate the potential of neural networks for the incremental \ppm scenario.
The goal is reached by (i) identifying different settings related to the the data to use for training, updating, and testing the models, both in a static and a dynamic scenario; and (ii) showing the positive impact of catastrophic forgetting of deep learning models for the \ppm use-case.
In our work, we focus on another type of techniques/predictions, i.e., we aim at investigating the potential of classical Machine Learning models in the \ppm scenario for the prediction of an outcome.

The work by~\citet{DBLP:conf/IEEEscc/MaisenbacherW17} is the only one we are aware of that exploits classical incremental Machine Learning in the context of \ppm. The goal of that paper is to show the usefulness of incremental techniques in the presence of Concept Drift. The goal is proved by performing an evaluation over synthetic logs which exhibit different types of Concept Drifts.
In our work, we aim at comparatively investigating four different model update strategies (which include the case of the incremental update) both in terms of accuracy of the results and in terms of time required to update the models. We carry on our evaluation on real-life and synthetic logs with and without an explicit Concept Drift.


\section{Conclusion}
\label{sec:conclusions}

In this paper, we have provided a first investigation of different update strategies for \ppm models in the context of the outcome prediction problem.
In particular, we have evaluated the performance of four update strategies, namely \textbf{do nothing}, \textbf{re-train with no hyperopt}, \textbf{full re-train}, and \textbf{incremental update}, applied to Random Forest, the reference technique for outcome-oriented predictions, on a number of real and synthetic \dss with and without explicit Concept Drift. The cost-effectiveness of the different update strategies has been evaluated in the simple case where only one train and one test set are available, and in the more complex scenario where new batches of data become continuously available.
The results show that the need to update a \ppm model is real for typical real-life event logs (regardless of the presence of an explicit Concept Drift). They also show the potential of incremental learning strategies for \ppm in real environments.
An avenue for future work is the extension of our evaluation to different prediction problems such as remaining time and sequence predictions, which would, in turn, extend the evaluation to different reference Machine Learning techniques such as regression and LSTM, respectively. Also, a deeper investigation of the proposed cost-effectiveness framework in the context of the proposed update strategies will allow us to come up with more detailed best practices to guide the user in understanding which strategy is the most appropriate one under specific contextual conditions.

To conclude, we believe that the potential of incremental models is under-appreciated in the \ppm field. To allow researchers to better understand the usefulness of the update strategies proposed in this paper, we made them readily available in the latest release of \nirdi \citep{DBLP:conf/bpm/RizziSFGKM19}.

\appendix

\section{Further Results}
\label{sec:appendix}

We report, in this appendix, the results obtained using Perceptron~\citep{perceptron} (rather than Random Forest) as classifier.
Tables~\ref{tab:perceptron_accuracy_a} and~\ref{tab:perceptron_accuracy_b} show the results obtained with such a classifier for the different strategies and for the different considered settings. Also for these experiments, the best result for each dataset and labelling is emphasised in italic, while the results that differ from the best ones for less than 0.01 are emphasised in bold.

As for the case of Random Forest, we can observe that the most accurate strategies are $\Mtwo$ and $\Mthree$, with the exception of datasets BPIC12 and BPIC15 with labelling $\phi_{32}$ in the setting 40\%-40\%, in which the best results are obtained with $\Mzero$.
Moreover, we can observe a significant difference, in terms of performance, with respect to the Random Forest, for BPIC18: the accuracy obtained with Perceptron for this dataset is very low for all the four update strategies.
Also with Perceptron, the accuracy is overall better for the 40\%-40\% setting w.r.t.\ 10\%-70\% and no relevant differences in terms of winning strategies for datasets with and without explicit Concept Drift can be devised.

Overall, the evaluation with Perceptron confirms that $\Mtwo$ and $\Mthree$ (i.e., \textbf{full re-train} and \textbf{incremental update}, respectively) are the best performing update strategies in terms of accuracy.

\begin{table}
	\centering
    \begin{subtable}{.5\linewidth}
		\centering
		\scalebox{.70}{
        \begin{tabular}{@{}c c c c c c@{}}
			\toprule
			\textbf{Dataset} & $ \mathbf{\phi} $ & $\mathbf{\Mzero}$ & $\mathbf{\Mone}$ & $\mathbf{\Mtwo}$ & $\mathbf{\Mthree}$ \\
			\midrule
			\multirow{3}*{BPIC11}   & $\phi _{11}$ & 0.504 & 0.536 & \textit{ \textbf{0.625}} & 0.566 \\
								    & $\phi _{12}$ & 0.682 & 0.413 & \textit{ \textbf{0.867}} & 0.749 \\
								    & $\phi _{13}$ & 0.676 & \textbf{0.877} & \textit{ \textbf{0.88}} &	0.863  \\
			\cmidrule{2-6}
			\multirow{3}*{BPIC12}   & $\phi _{21}$ & \textit{ \textbf{0.69}}	& 0.633 & 0.655 &	0.68 \\
								    & $\phi _{22}$ & \textit{ \textbf{0.72}} &	\textbf{0.711}	& 0.686	& \textbf{0.719} \\
								    & $\phi _{23}$ & \textbf{0.515} & \textit{ \textbf{0.518}} & 0.483 & 0.5 \\
			\cmidrule{2-6}
			\multirow{3}*{BPIC15}   & $\phi _{31}$ & 0.675 & 0.725 & 0.732 & \textit{ \textbf{0.777}} \\
								    & $\phi _{32}$ & 0.827 & 0.809 & 0.832 & \textit{ \textbf{0.873}}  \\
								    & $\phi _{33}$ & 0.745 & 0.841 & \textit{ \textbf{0.886}} & 0.788 \\
			\cmidrule{2-6}
			BPIC18   & $\phi _{41}$ & 0.386	& 0.392 &  0.397 &	\textit{ \textbf{0.509}} \\
			\cmidrule{2-6}
			DriftRIO1 & $\phi _{51}$ & 0.369 & \textbf{0.738} & \textbf{0.742} & \textit{ \textbf{0.747}} \\
			\cmidrule{2-6}
			DriftRIO2	& $\phi _{61}$ & 0.701 & 0.945 & \textit{ \textbf{0.961}} & 0.908 \\
			\bottomrule
        \end{tabular}
		}
		\caption{Setting 10\%-70\%.}
		\label{tab:perceptron_accuracy_a}
    \end{subtable}%
    \begin{subtable}{.5\linewidth}
		\centering
		\scalebox{.70}{
        \begin{tabular}{@{}c c c c c c@{}}
			\toprule
			\textbf{Dataset} & $ \mathbf{\phi} $ & $\mathbf{\Mzero}$ & $\mathbf{\Mone}$ & $\mathbf{\Mtwo}$ & $\mathbf{\Mthree}$ \\
			\midrule
			\multirow{3}*{BPIC11}   & $\phi _{11}$ & 0.596 & 0.643 & 0.625 & \textit{ \textbf{0.666}}  \\
								    & $\phi _{12}$ & 0.802 & 0.788 & \textit{ \textbf{0.867}} & 0.807 \\
								    & $\phi _{13}$ & 0.826 & \textbf{0.873}& \textit{ \textbf{0.88}} & \textbf{0.879} \\
			\cmidrule{2-6}
			\multirow{3}*{BPIC12}   & $\phi _{21}$ & \textit{ \textbf{0.658}} & 0.646 & \textbf{0.655} & 0.478 \\
								    & $\phi _{22}$ & \textbf{0.682} & \textit{ \textbf{0.69}} &	\textbf{0.686}	& 0.574 \\
								    & $\phi _{23}$ & \textit{ \textbf{0.524}} &	0.501 & 0.483 &	\textbf{0.521} \\
			\cmidrule{2-6}
			\multirow{3}*{BPIC15}   & $\phi _{31}$ & 0.651 & 0.698 & \textbf{0.732} & \textit{ \textbf{0.736}}  \\
								    & $\phi _{32}$ & \textit{ \textbf{0.854}} & 0.801 & 0.832 & 0.836 \\
								    & $\phi _{33}$ & \textbf{0.88} &	0.864 &	\textit{ \textbf{0.886}} &	0.876  \\
			\cmidrule{2-6}
			BPIC18        & $\phi _{41}$ & 0.426 & 0.446 & 0.397 & 	\textit{ \textbf{0.574}} \\
			\cmidrule{2-6}
			DriftRIO1 & $\phi _{51}$ & 0.727 & 0.747 & 0.742 &	\textit{ \textbf{0.908}} \\
			\cmidrule{2-6}
			DriftRIO2 & $\phi _{61}$ & 0.594 & 0.939 & \textit{ \textbf{0.961}} & 0.839 \\
			\bottomrule
        \end{tabular}
		}
		\caption{Setting 40\%-40\%.}
		\label{tab:perceptron_accuracy_b}
	\end{subtable}
	\caption{The accuracy results related to the Perceptron model.}
	\label{tab:perceptron_accuracy}	
\end{table}

\bibliographystyle{plain}

\end{document}